\documentclass[letterpaper]{article} % DO NOT CHANGE THIS
\usepackage{aaai23}  % DO NOT CHANGE THIS
\usepackage{times}  % DO NOT CHANGE THIS
\usepackage{helvet}  % DO NOT CHANGE THIS
\usepackage{courier}  % DO NOT CHANGE THIS
\usepackage[hyphens]{url}  % DO NOT CHANGE THIS
\usepackage{graphicx} % DO NOT CHANGE THIS
\usepackage{natbib}  % DO NOT CHANGE THIS AND DO NOT ADD ANY OPTIONS TO IT
\usepackage{caption} % DO NOT CHANGE THIS AND DO NOT ADD ANY OPTIONS TO IT
\usepackage{algorithm}
\usepackage{algpseudocode}
\usepackage[hidelinks]{hyperref}

\usepackage{newfloat}
\usepackage{listings}
\DeclareCaptionStyle{ruled}{labelfont=normalfont,labelsep=colon,strut=off} % DO NOT CHANGE THIS
\floatstyle{ruled}
\newfloat{listing}{tb}{lst}{}
\floatname{listing}{Listing}
\usepackage{booktabs}
\usepackage{soul}
\usepackage{cuted}
\usepackage{caption}
\usepackage{subcaption}
\usepackage[table, dvipsnames]{xcolor}
\usepackage{multirow}

\usepackage{times}
\usepackage{graphicx}
\usepackage{amsmath}
\usepackage{amssymb}
\usepackage[accsupp]{axessibility}  % Improves PDF readability for those with disabilities.

\usepackage{xspace}
\DeclareMathOperator*{\argmin}{arg\,min} % thin space, limits underneath in displays

\makeatletter

\DeclareRobustCommand\onedot{\futurelet\@let@token\@onedot}
\def\@onedot{\ifx\@let@token.\else.\null\fi\xspace}

\def\eg{\emph{e.g}\onedot} 
\def\ie{\emph{i.e}\onedot}

\def\wrt{w.r.t\onedot} 
\def\etal{\emph{et al}\onedot}
\makeatother

\title{
Dropout is NOT All You Need to Prevent Gradient Leakage
}
\author{
    Daniel Scheliga\textsuperscript{\rm 1},
    Patrick M\"{a}der\textsuperscript{\rm 1,2},
    Marco Seeland\textsuperscript{\rm 1}
}
\affiliations {
    \textsuperscript{\rm 1} Technische Universit\"{a}t Ilmenau, Germany\\
    \textsuperscript{\rm 2} Friedrich Schiller Universit\"{a}t Jena, Germany\\
    daniel.scheliga@tu-ilmenau.de, patrick.maeder@tu-ilmenau.de, marco.seeland@tu-ilmenau.de
}

\begin{document}
\maketitle

%%%%%%%%% ABSTRACT
\begin{abstract}
% Context and motivation
Gradient inversion attacks on federated learning systems reconstruct client training data from exchanged gradient information.
To defend against such attacks, a variety of defense mechanisms were proposed.
However, they usually lead to an unacceptable trade-off between privacy and model utility.
% Question / Problem
Recent observations suggest that dropout could mitigate gradient leakage and improve model utility if added to neural networks.
Unfortunately, this phenomenon has not been systematically researched yet.
% Principal ideas / results
In this work, we thoroughly analyze the effect of dropout on iterative gradient inversion attacks.
We find that state of the art attacks are not able to reconstruct the client data due to the stochasticity induced by dropout during model training.
Nonetheless, we argue that dropout does not offer reliable protection if the dropout induced stochasticity is adequately modeled during attack optimization.
Consequently, we propose a novel \textit{Dropout Inversion Attack (DIA)} that jointly optimizes for client data and dropout masks to approximate the stochastic client model.
% Contribution
We conduct an extensive systematic evaluation of our attack on four seminal model architectures and three image classification datasets of increasing complexity.
We find that our proposed attack bypasses the protection seemingly induced by dropout and reconstructs client data with high fidelity.
Our work demonstrates that privacy inducing changes to model architectures alone cannot be assumed to reliably protect from gradient leakage and therefore should be combined with complementary defense mechanisms.
\end{abstract}

\section{Introduction}
\label{sec:intro}
Federated Learning strategies were designed to leverage the collaborative use of distributed data to learn a common machine learning model.
Since training data is not shared between participating clients, systemic privacy risks can be mitigated~\cite{Kairouz2019}.
Recent work, however, shows that the privacy of participating clients can be compromised by reconstructing sensitive data from gradients or model states that are exchanged during the federated training.
The most versatile reconstruction techniques are realized as iterative gradient inversion attacks~\cite{Zhu2019, Zhao2020, Wei2020, Geiping2020, Yin2021, Lu2021, Hatamizadeh2022}.
These attacks optimize randomly initialized dummy images so that their resulting dummy gradients match the targeted client gradient.

\begin{figure}
     \centering
     \begin{minipage}[c]{.24\linewidth}
     \centering Original\\
     \begin{subfigure}{.96\linewidth}
         \centering
         \vspace{8pt}
         \includegraphics[width=.95\linewidth]{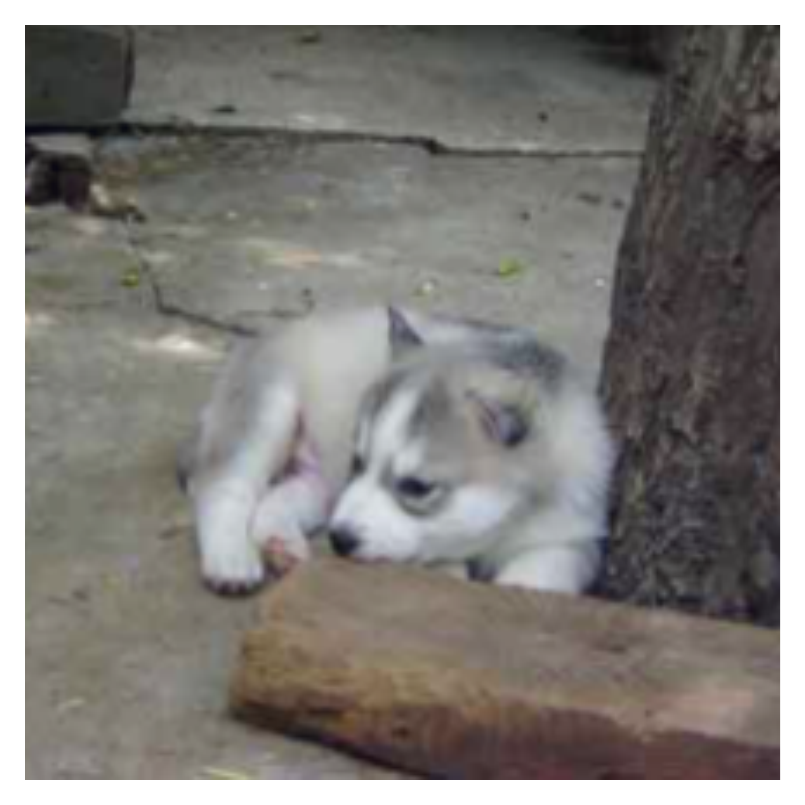}  
         \caption{}
     \end{subfigure}
     \end{minipage}
     \begin{minipage}[c]{.24\linewidth}
        \centering
        No Dropout\\ 
     \begin{subfigure}{.96\linewidth}
         \centering
         IG
         \includegraphics[width=.95\linewidth]{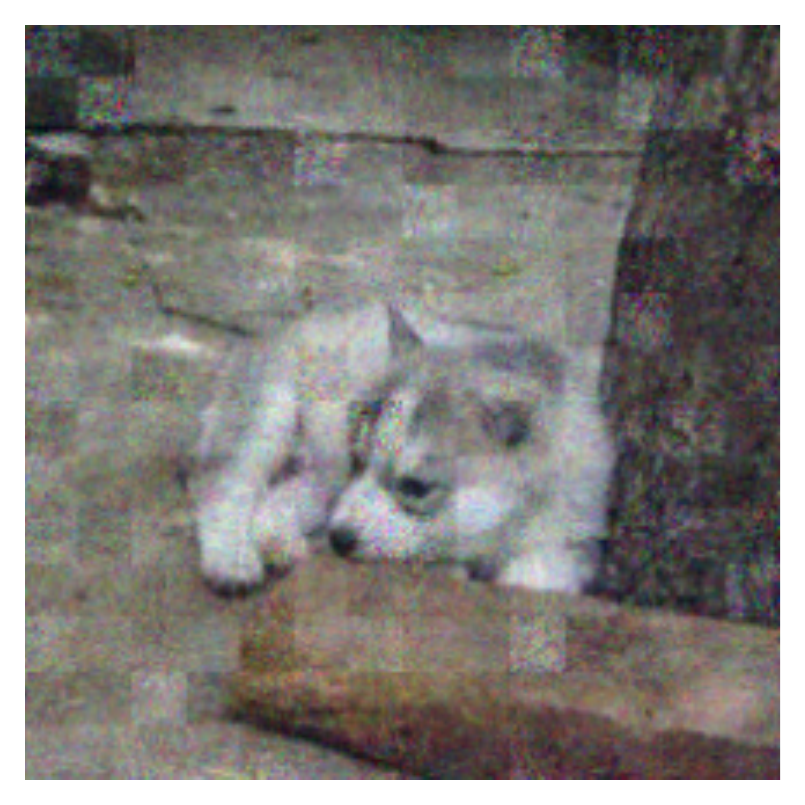}  
         \caption{}
     \end{subfigure}
     \end{minipage}
     \begin{minipage}[c]{.48\linewidth}
     \centering
     Dropout ($p=0.1$)\\
     \begin{subfigure}{.48\linewidth}
         \centering
         IG
         \includegraphics[width=.95\linewidth]{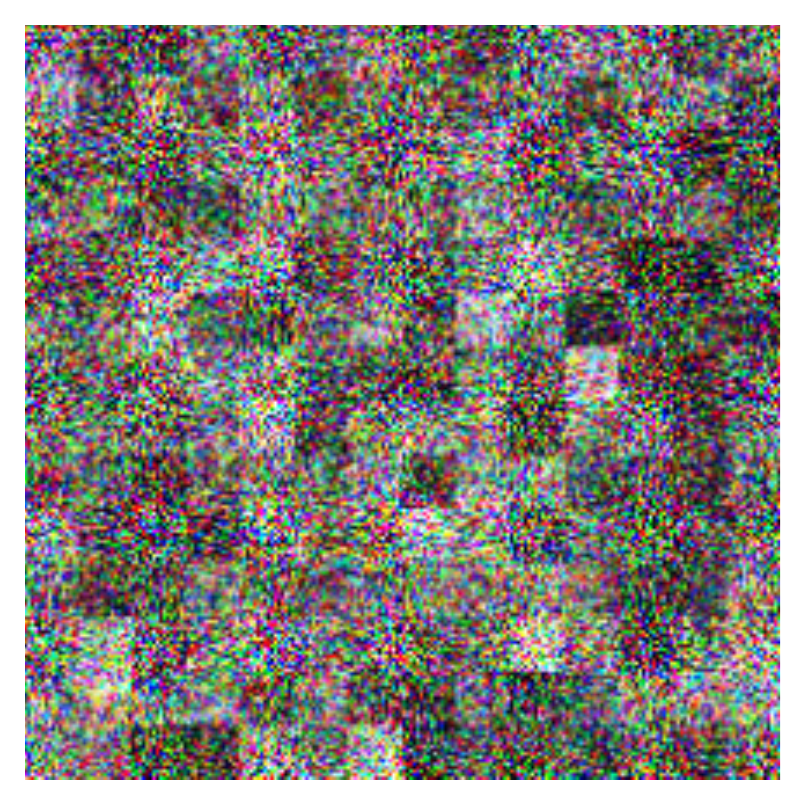}  
         \caption{}
     \end{subfigure}
     \begin{subfigure}{.48\linewidth}
         \centering
         Ours
         \includegraphics[width=.95\linewidth]{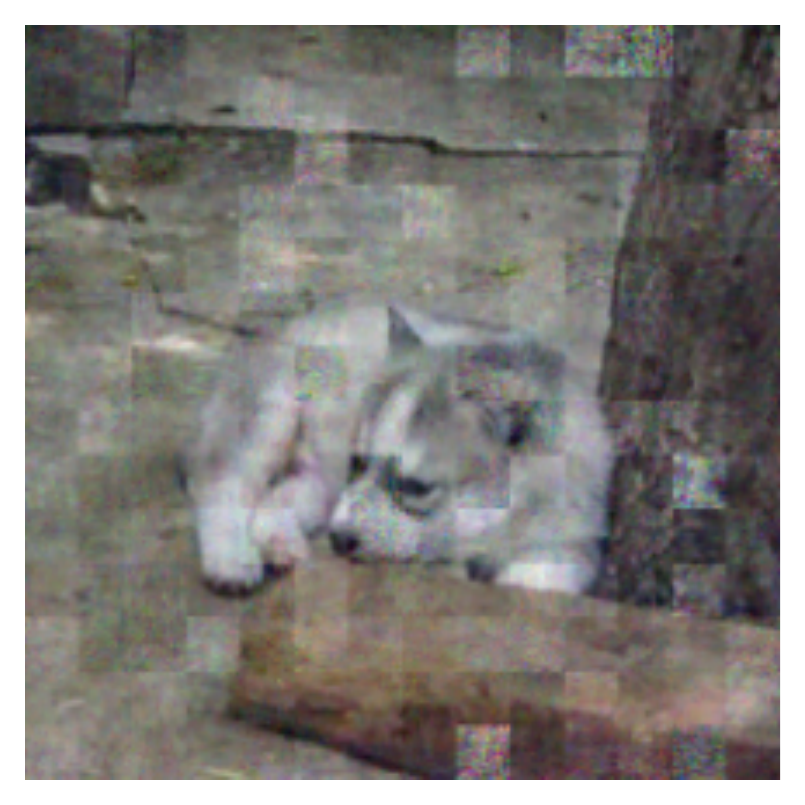}  
         \caption{}
     \end{subfigure}
     \end{minipage}
     \caption{\textbf{Reconstructing data from gradients without and with dropout.} (a) Original image. (b) State of the art IG attack~\cite{Geiping2020} without dropout. (c) State of the art IG attack~\cite{Geiping2020} with dropout. (d) Our proposed Dropout Inversion Attack with dropout.}
     \label{fig:teaser}
\end{figure}

Defense strategies to protect against such attacks are based on: (1) adjustments to the training process, e.g. increasing the number of local training iterations or the batchsize~\cite{Wei2020}, (2) changes to the input data, e.g. perturbation or input encryption~\cite{Huang2020a, Huang2020b}, (3) perturbation of exchanged gradient information, e.g. through the addition of noise, compression or pruning~\cite{Bonawitz2017, Jayaraman2019, Zhu2019, Papernot2019, Sattler2020, Lyu2021, Wei2021b}, or (4) application of specifically designed architectural features or modules~\cite{Scheliga2022a, Scheliga2022b, Sun2021}.
The use of most defense mechanisms, however, results in a trade-off between privacy and model utility~\cite{Dwork2013a, Jayaraman2019, Zhu2019, Wei2020, Huang2021, Scheliga2022a, Scheliga2022b}.

Dropout is a regularization technique that aims to reduce overfitting in deep neural networks~\cite{Hanson1990,Hinton2012}.
While the use of dropout can boost the performance of neural networks~\cite{Srivastava2014}, recent publications suggest that it could also protect shared gradients from gradient leakage~\cite{Wei2020, Zheng2021}.
Inspired by these observations, we show that the stochasticity introduced by dropout indeed protects shared gradients from gradient leakage through iterative gradient inversion attacks.
However, we claim that this protection is only apparent, because the attacker has no access to the specific realization of the stochastic client model used during training.
Moreover, we argue that an attacker can sufficiently approximate this specific realization of the client model using the shared gradient information.
To reveal the vulnerability of dropout protected models, we formulate a novel \textbf{Dropout Inversion Attack} (DIA) that jointly optimizes for client data and the dropout masks applied during local training.

Our contributions can be summarized as follows:
\begin{itemize}
    \item We systematically show that the application of dropout during neural network training seems to prevent gradient leakage by iterative gradient inversion attacks.
    \item We formulate a novel attack that, contrary to previous attacks, successfully reconstructs client training data from dropout protected shared gradients.
    Note that the components of our proposed attack can be universally used to extend any other iterative gradient inversion attack.
    \item We perform an extensive systematic evaluation of our attack on two dense connection based (Multi Layer Perceptron, Vision Transformer) and two CNN based (LeNet, ResNet) model architectures as well as three image classification datasets of increasing complexity (MNIST, CIFAR-10, ImageNet).
\end{itemize}

\section{Related Work}
\label{sec:sota}
\subsection{Gradient Inversion Attacks}
\label{sec:sotF_gia}

Consistent with related work~\cite{Geiping2020, Enthoven2020a, Yin2021, Kaissis2021, Jin2021, Scheliga2022a, Scheliga2022b, Zhang2022, Gupta2022}, we assume a \textit{honest-but-curious} server threat model.
In this scenario the attacker has insight into the training process, \ie knowledge of the model $F$, the loss function $\mathcal{L}$ used to optimize the model parameters $\theta$ and the client gradient $\nabla \mathcal{L}_\theta(F(x), y)$ which is exchanged during federated training.
Given this knowledge, the attacker aims to reconstruct training data $(x,y)$ of clients that participate in federated training.

To achieve this, the attacker iteratively minimizes the distance $D$ between the client gradient $\nabla \mathcal{L}_\theta(F(x), y)$ and a dummy gradient $\nabla \mathcal{L}_\theta(F(x'),y')$.
The dummy gradient is obtained by forward propagation of randomly initialized dummy data $(x', y')$ through the model $F$.
A gradient based optimizer,~\eg Adam~\cite{Kingma2014}, adjusts the dummy data $(x', y')$ until convergence.

Attack optimization can be formally expressed as:
\begin{equation}
    \label{eq:GI}
    \argmin_{ \left( x',y' \right) } D(\nabla \mathcal{L}_\theta(F(x), y),  \nabla \mathcal{L}_\theta(F(x'),y')) + \lambda \Omega.
\end{equation}
Depending on the specific attack, the regularization term $\lambda \Omega$ can take different forms. 
Generally $\lambda \Omega$ aims to stabilize optimization and to improve reconstruction quality.

The first iterative gradient inversion attack was introduced by Zhu \etal~\cite{Zhu2019}.
They use Euclidean distance for $D$ and no regularization.
The authors of~\cite{Zhao2020} and~\cite{Yin2021} proposed methods to analytically reconstruct the ground-truth labels $y$ in advance.
As long as the training batch contains disjoint classes, an attacker can reliably reconstruct label information.
This eliminates optimization for $y'$ in Eq.~\ref{eq:GI} and accelerates the overall attack.

Geiping~\etal further improve the reconstruction process with their \textit{Inverting Gradients} (IG) attack~\cite{Geiping2020}.
They minimize the cosine distance between client and dummy gradients instead of Euclidean distance to disentangle gradient direction and magnitude.
Furthermore, they add a total variation~\cite{Rudin1992} prior of the dummy image $x'$ as regularization term to increase the fidelity of their reconstructions.

Lu~\etal~\cite{Lu2021} specifically target transformer based architectures. 
They find that the trainable position embedding in transformers can be greatly abused for reconstruction.
Their iterative \textit{Attention PRIvacy Leakage} (APRIL) attack uses Euclidean distance for $D$ and adds the cosine distance between client and dummy gradients of the positional embedding as regularization term.

Other related work in the area of iterative gradient inversion attacks mainly focuses to improve the reconstruction quality through the choice of the 1) gradient distance function $D$, 2) regularization term $\Omega$, 3) initialization of the dummies $(x', y')$ and 4) label reconstruction method~\cite{Wei2020, Wang2020c, Yin2021, Jin2021, Jeon2021}.
A detailed overview of recent attack combinations can be found in \cite{Li2022} and \cite{Zhang2022}.

\subsection{Dropout}
\label{sec:sotF_do}
Dropout~\cite{Hanson1990, Hinton2012} is a commonly used regularization method that randomly masks the output of neurons with a chosen probability $p$.
Hence, each forward pass realizes a different version of the neural network.
This makes dropout an efficient technique for model averaging and in turn prevents models from overfitting to training data~\cite{Srivastava2014}.

Formally, we consider a neural network $F: X \rightarrow Y$, $F(x)=y$ to be a deterministic function that calculates an output $y \in Y$ from an input $x \in X$.
Given the output $z^{(i)}$ of the $i$th layer $L^{(i)}$ in $F$, a succeeding dropout layer $L^{(i)}_D$ multiplies $z^{(i)}$ element-wise with a random dropout mask $\psi^{(i)}$ and scales the remaining outputs according to the dropout rate $p$ to preserve the output magnitude: 
\begin{equation}
    \label{eq:dropout}
    L^{(i)}_D(z^{(i)}) = \frac{1}{1-p} \cdot z^{(i)} \circ \psi^{(i)}
\end{equation}
For every dropout layer $L^{(i)}_D$ $i\in \lbrace 1,\dots,l \rbrace$, $\psi^{(i)}$ is a vector of independent Bernoulli variables, \ie $\psi^{(i)} \sim \text{Bernoulli}(p)$.
We define $\tilde\Psi_p = \lbrace \psi^{(1)},\dots,\psi^{(l)} \rbrace$ as the set of $l$ random dropout masks for a neural network with $l$ dropout layers.

The use of dropout turns a deterministic neural network into a stochastic one.
Hence, the set of all functions $F$ that depend on the dropout masks $\tilde\Psi_p$ is $\mathcal{F}_p = \lbrace F_\Psi | \Psi\sim\tilde\Psi_p\rbrace$.
We denote $\Psi$ as one arbitrary but fixed sample from $\tilde\Psi_p$.
At each training step a new $\Psi$ is sampled. 
Consequently, this realizes a different version $F_\Psi \in \mathcal{F}_p$ of the neural network that is used for forward propagation and gradient calculation.

As dropout introduces noise into the training process, a decrease in reconstruction quality of iterative gradient inversion attacks is observed in recent work~\cite{Wei2020, Zheng2021}.
Contrary to these findings, Enthoven~\etal~\cite{Enthoven2020a} find that the use of dropout after the first fully connected layer of a neural network increases the success to analytically reconstruct client data from larger batches. 
Such analytical attacks, however, can be easily mitigated by removing bias weights from the model~\cite{Scheliga2022b}.

\section{Dropout vs Gradient Leakage}
\label{sec:dvpl}
Although no systematic studies have yet been conducted, recent observations suggest that dropout can decrease the success of iterative gradient inversion attacks~\cite{Wei2020,Zheng2021}. 
To confirm these observations, we first conduct a series of experiments that evaluate the effect of increased dropout rates on reconstruction quality and model utility.
Next, we argue that an attacker would be able to successfully reconstruct client training data if given knowledge about the specific realization of the stochastic client model.
Therefore, we conduct proof of concept experiments that consider an attacker who knows the dropout masks applied during client model training,~\ie a \textit{well-informed attacker}.

\subsection{Attacking dropout protected models}
\label{sec:dvpl_baseline}
To confirm the impact of dropout on iterative gradient inversion attacks, we first attack a Multi Layer Perceptron (MLP)~\cite{Rumelhart1985} and a Vision Transformer (ViT)~\cite{Dosovitskiy2020} trained on the MNIST~\cite{Deng2012} and CIFAR-10~\cite{Krizhevsky2009} datasets.
We chose these architectures as they typically use dropout as regularization technique.
We use the publicly available PyTorch implementation of IG\footnote{\url{https://github.com/JonasGeiping/invertinggradients}} provided by~\cite{Geiping2020} as gradient inversion attack.

To observe the effect of dropout on model utility we follow the federated scenario and hyperparameters used in~\cite{Scheliga2022b}.
We report the test accuracy of the global model state after convergence.
More details on the experimental setup can be found in Section~\ref{sec:exp}.

When dropout is used, the attacker has two options to generate the dummy gradients required for attack optimization.
Analogous to client training, the first option uses the model in training mode,~\ie the stochastic model.
In this case, the attacker applies randomly sampled dropout masks $\Psi_A$ in each forward propagation so that a different realization $F_{\Psi_A} \in\mathcal{F}_p$ is used in each iteration during attack optimization.
Consequently, the dummy gradients $\nabla \mathcal{L}_\theta(F_{\Psi_A}(x'),y'))$ differ greatly for each attack iteration and are "elusive and unable to converge"~\cite{Wei2020} to match the client gradient $\nabla \mathcal{L}_\theta(F_{\Psi_C}(x),y))$.

The second option uses the model in inference mode,~\ie dropout is not applied.
Note that in this case all dropout masks $\psi_A^{(i)} = I$ $\forall i=1,\dots,l$.
Hence, the same realization $F_{\Psi_A}\in \mathcal{F}_{p=0} = \lbrace F \rbrace$ is used in each iteration during attack optimization.
The attacker's dummy gradients are more stable compared to when the stochastic model is used.
However, since the client used dropout during training, $\Psi_A \neq \Psi_C$ causes the dummy gradients to differ from the client gradients despite attack optimization.

\begin{table}[t]
\centering
\resizebox{0.95\linewidth}{!}{%
\begin{tabular}{c|c|c|c||c|c}
\toprule
 & & & & IG & WIIG \\
 & Model & $p$ & Accuracy [\%] $\uparrow$ & SSIM $\uparrow$ & SSIM $\uparrow$ \\
\midrule
\multirow[c]{8}{*}{\rotatebox{90}{MNIST}} & \multirow[c]{4}{*}{MLP} & 0.00& {\cellcolor[HTML]{097940}} \color[HTML]{F1F1F1} 98.53& {\cellcolor[HTML]{006837}} \color[HTML]{F1F1F1} \bfseries 1.00& {\cellcolor[HTML]{FFFFFF}} \color[HTML]{000000} - \\
 &  & 0.25& {\cellcolor[HTML]{0F8446}} \color[HTML]{F1F1F1} 98.28& {\cellcolor[HTML]{6BBF64}} \color[HTML]{000000} 0.79& {\cellcolor[HTML]{006837}} \color[HTML]{F1F1F1} \bfseries 1.00\\
 &  & 0.50& {\cellcolor[HTML]{30A356}} \color[HTML]{F1F1F1} 97.48& {\cellcolor[HTML]{DCF08F}} \color[HTML]{000000} 0.59& {\cellcolor[HTML]{006837}} \color[HTML]{F1F1F1} \bfseries 1.00\\
 &  & 0.75& {\cellcolor[HTML]{F4FAB0}} \color[HTML]{000000} 93.50& {\cellcolor[HTML]{FDAD60}} \color[HTML]{000000} 0.30& {\cellcolor[HTML]{57B65F}} \color[HTML]{F1F1F1} \itshape 0.82\\
\cline{2-6}
 & \multirow[c]{4}{*}{ViT} & 0.00& {\cellcolor[HTML]{04703B}} \color[HTML]{F1F1F1} 98.76& {\cellcolor[HTML]{05713C}} \color[HTML]{F1F1F1} 0.98& {\cellcolor[HTML]{FFFFFF}} \color[HTML]{000000} - \\
 &  & 0.25& {\cellcolor[HTML]{006837}} \color[HTML]{F1F1F1} \bfseries 98.98& {\cellcolor[HTML]{B91326}} \color[HTML]{F1F1F1} 0.04& {\cellcolor[HTML]{026C39}} \color[HTML]{F1F1F1} 0.99\\
 &  & 0.50& {\cellcolor[HTML]{06733D}} \color[HTML]{F1F1F1} 98.67& {\cellcolor[HTML]{AF0926}} \color[HTML]{F1F1F1} \itshape 0.02& {\cellcolor[HTML]{026C39}} \color[HTML]{F1F1F1} 0.99\\
 &  & 0.75& {\cellcolor[HTML]{A50026}} \color[HTML]{F1F1F1} \itshape 87.36& {\cellcolor[HTML]{AF0926}} \color[HTML]{F1F1F1} \itshape 0.02& {\cellcolor[HTML]{006837}} \color[HTML]{F1F1F1} \bfseries 1.00\\
\midrule
\multirow[c]{8}{*}{\rotatebox{90}{CIFAR-10}} & \multirow[c]{4}{*}{MLP} & 0.00& {\cellcolor[HTML]{C9E881}} \color[HTML]{000000} 54.72& {\cellcolor[HTML]{006837}} \color[HTML]{F1F1F1} \bfseries 1.00& {\cellcolor[HTML]{FFFFFF}} \color[HTML]{000000} - \\
 &  & 0.25& {\cellcolor[HTML]{E0F295}} \color[HTML]{000000} 52.52& {\cellcolor[HTML]{AFDD70}} \color[HTML]{000000} 0.68& {\cellcolor[HTML]{05713C}} \color[HTML]{F1F1F1} \bfseries 0.98\\
 &  & 0.50& {\cellcolor[HTML]{FA9B58}} \color[HTML]{000000} 38.89& {\cellcolor[HTML]{FBFDBA}} \color[HTML]{000000} 0.51& {\cellcolor[HTML]{45AD5B}} \color[HTML]{F1F1F1} 0.84\\
 &  & 0.75& {\cellcolor[HTML]{A50026}} \color[HTML]{F1F1F1} \itshape 27.09& {\cellcolor[HTML]{FED683}} \color[HTML]{000000} 0.38& {\cellcolor[HTML]{73C264}} \color[HTML]{000000} \itshape 0.78\\
\cline{2-6}
 & \multirow[c]{4}{*}{ViT} & 0.00& {\cellcolor[HTML]{3CA959}} \color[HTML]{F1F1F1} 64.47& {\cellcolor[HTML]{30A356}} \color[HTML]{F1F1F1} 0.87& {\cellcolor[HTML]{FFFFFF}} \color[HTML]{000000} - \\
 &  & 0.25& {\cellcolor[HTML]{006837}} \color[HTML]{F1F1F1} \bfseries 70.83& {\cellcolor[HTML]{A90426}} \color[HTML]{F1F1F1} 0.01& {\cellcolor[HTML]{118848}} \color[HTML]{F1F1F1} 0.93\\
 &  & 0.50& {\cellcolor[HTML]{16914D}} \color[HTML]{F1F1F1} 67.01& {\cellcolor[HTML]{A90426}} \color[HTML]{F1F1F1} 0.01& {\cellcolor[HTML]{0A7B41}} \color[HTML]{F1F1F1} 0.96\\
 &  & 0.75& {\cellcolor[HTML]{FEE491}} \color[HTML]{000000} 45.08& {\cellcolor[HTML]{A50026}} \color[HTML]{F1F1F1} \itshape 0.00& {\cellcolor[HTML]{0C7F43}} \color[HTML]{F1F1F1} 0.95\\
\bottomrule
\end{tabular}
}
\caption{
%Model accuracy after federated training and SSIM for a MLP and ViT on MNIST and CIFAR-10. 
%Gradients are attacked with IG~\cite{Geiping2020}. 
Model accuracy after federated training of a MLP and ViT on MNIST and CIFAR-10 as well as SSIM computed from gradients attacked with IG. 
WIIG indicates that the attacker has knowledge of the victim dropout masks $\Psi_V$.
%WIIG indicates a well-informed attacker that has knowledge of the victim dropout masks $\Psi_V$.
Arrows indicate direction of improvement.
Bold and italic formatting highlight best and worst results respectively.
}
\label{tab:IGAPM}
\end{table}

Tab.~\ref{tab:IGAPM} shows the global model accuracy after federated training of the MLP and ViT on MNIST and CIFAR-10, as well as the privacy as measured by SSIM.
Dropout rates were selected as $p\in \lbrace 0, 0.25, 0.50, 0.75 \rbrace$.
With increasing $p$ the SSIM steadily decreases for the MLP; hence, privacy increases.
However, we also observe a negative impact of dropout on MLP model utility.
Findings in~\cite{Hofmann2021} confirm this effect.
Furthermore, Piotrowski~\etal~\cite{Piotrowski2020} argue that MLPs with a low width require very low dropout rates to achieve improvements in model utility.

\begin{figure}
     \centering
     \begin{subfigure}{.95\linewidth}
         \centering
         \includegraphics[width=\linewidth]{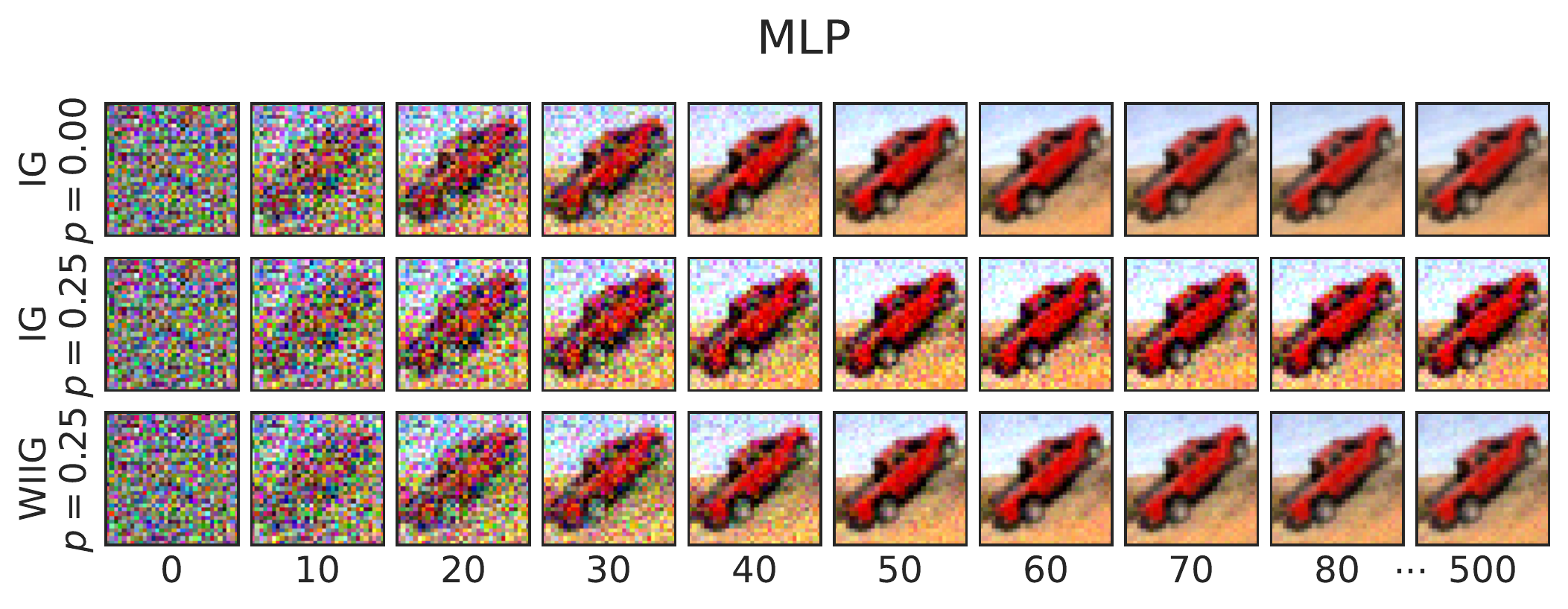}  
     \end{subfigure}
     \begin{subfigure}{.95\linewidth}
         \centering
         \includegraphics[width=\linewidth]{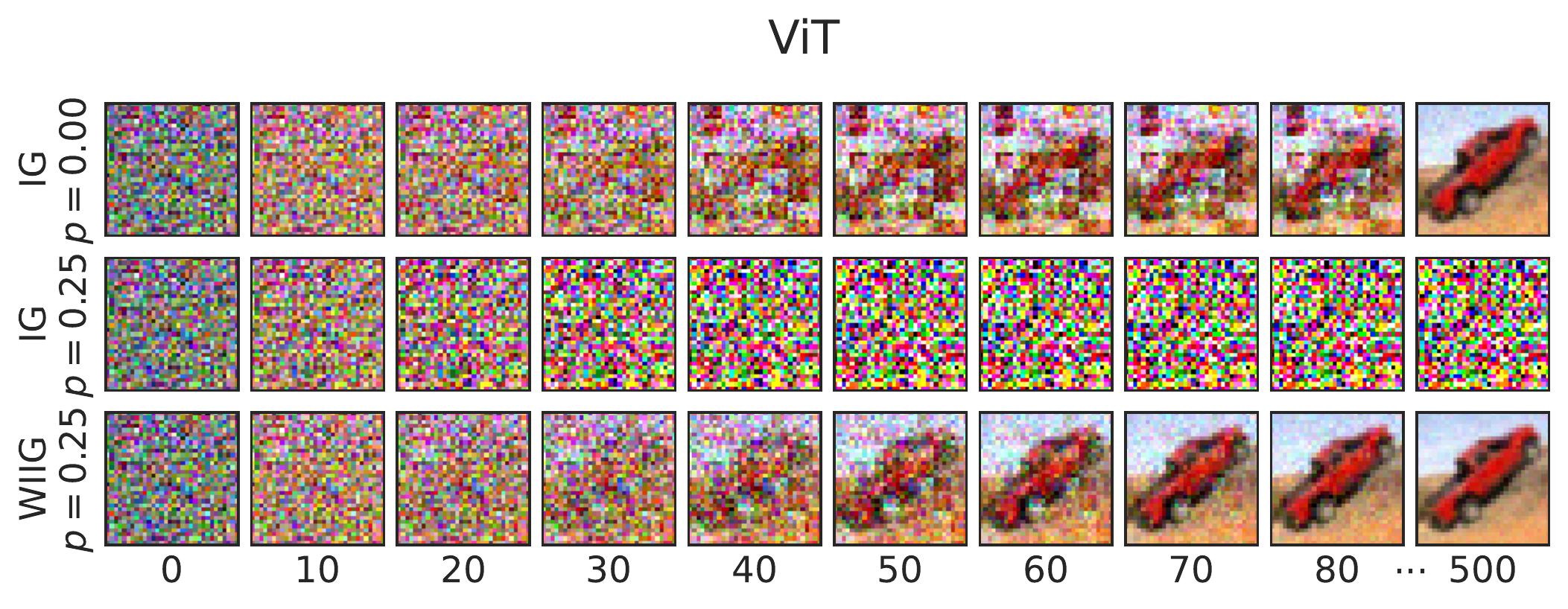}  
     \end{subfigure}
     \caption{
     \textbf{Exemplary reconstruction progress} for a MLP and ViT on CIFAR-10.
     WIIG indicates that the attacker has knowledge of the client dropout masks $\Psi_C$.
     Numbers on the ordinate indicate the attack iteration.
    }
    \label{fig:reconstruction_prog}
\end{figure}

The effect of dropout is even more pronounced for the ViT architecture.
A moderate dropout rate $p=0.25$ causes the SSIM to immediately drop from $0.98/0.87$ to $0.04/0.01$ for MNIST/CIFAR-10, respectively.
No visually recognizable information can be reconstructed (cf. Fig.~\ref{fig:reconstruction_prog} and~\ref{fig:mlpvit_rec}).
Furthermore, the accuracy of the ViT benefits from dropout with an absolute increase of $0.22\%/6.36\%$ for MNIST/CIFAR-10 at $p=0.25$.
Note that we have also used APRIL~\cite{Lu2021} to attack the ViT but found IG to perform better when dropout is applied.
More detailed results on the comparison of IG and APRIL, as well as more reconstruction quality metrics can be found in the technical appendix.
To ensure a consistent experimental setup, we stick with IG as baseline attack for the remaining experiments.

\begin{figure}
    \centering
    \includegraphics[width=\linewidth]{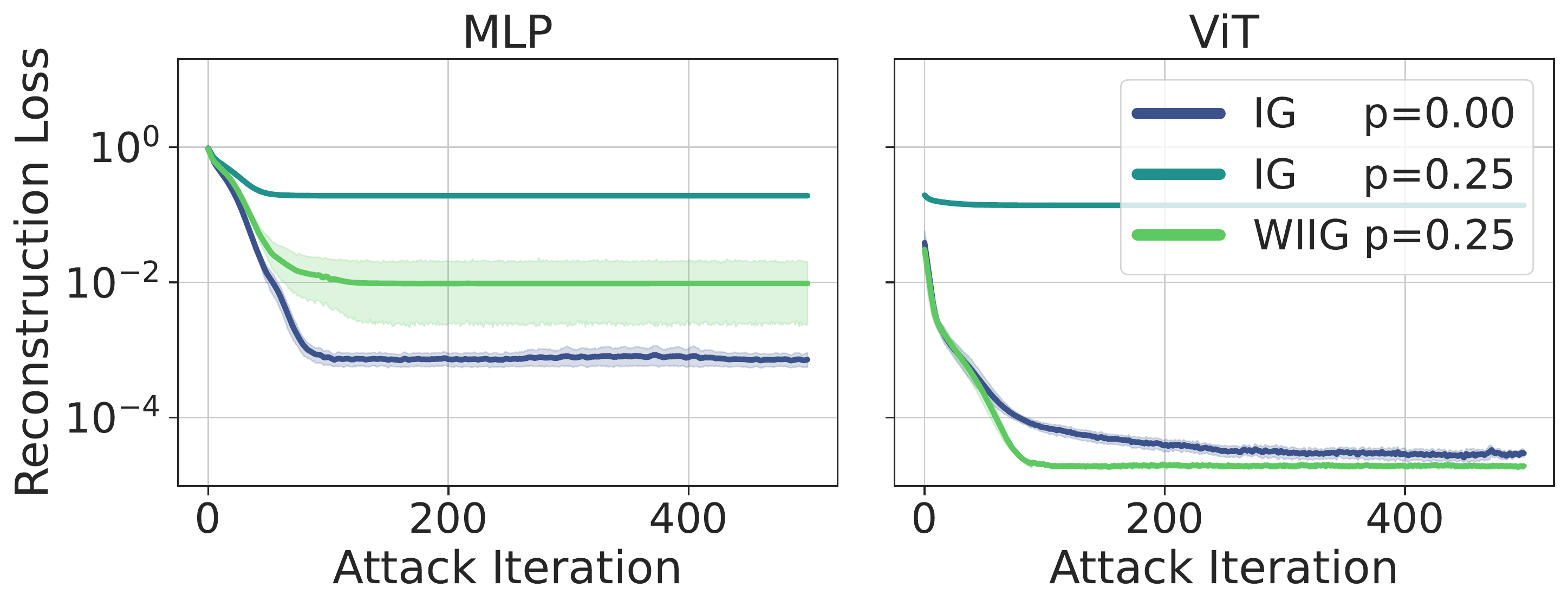}
    \caption{
    \textbf{Exemplary reconstruction loss} for a MLP and ViT on CIFAR-10. 
    WIIG indicates that the attacker has knowledge of the client dropout masks $\Psi_C$.
    }
    \label{fig:reconstruction_loss_pm}
\end{figure}

Fig.~\ref{fig:reconstruction_loss_pm} illustrates the behavior of the reconstruction loss during attack optimization.
Without dropout, \ie $p=0$ and hence $\Psi_A = \Psi_C$ (blue lines in Fig.~\ref{fig:reconstruction_loss_pm}), the dummy gradients quickly converge towards the client gradients.
The optimization becomes unstable as soon as dropout is used,~\eg with a dropout rate of $p=0.25$.
The attacker is forced to base the attack optimization on a model realization $F_{\Psi_A}$ that is different from the realization $F_{\Psi_C}$ used during training.
This causes a mismatch between dummy and client gradients.
Corresponding visual examples are displayed in Fig.~\ref{fig:reconstruction_prog}.

\subsection{The Well-Informed Attacker}
\label{sec:dvpl_pm}

The previous experiments show that the attack optimization cannot converge because the attacker and the client calculate their gradients based on different realizations $F_{\Psi_A}$ and $F_{\Psi_C}$.
We argue that the attacker would be able to reconstruct the client's training data if she is either informed about $F_{\Psi_C}$ or finds a suitable approximation thereof.
As a proof of concept, we conduct a series of experiments where the attacker applies the same dropout masks that were applied by the client during training,~\ie we use a \textit{well-informed attacker}.
Consequently, the attack optimization is based on the same realization $F_{\Psi_A} = F_{\Psi_C}$, and the gradient matching loss can be effectively minimized as in a model without dropout.

To empirically validate this argumentation we give the attacker knowledge over $\Psi_C$. 
During the iterative attack optimization, the attacker applies $\Psi_C$ when forward propagating the dummy images to calculate the dummy gradients.
We denote this as \textit{well-informed inverting gradients} attack, in short WIIG.
The remainder of the IG attack remains unchanged.

Tab.~\ref{tab:IGAPM} displays the reconstruction quality measured in SSIM for the well-informed attacker. 
The MLP still shows a slight decrease in SSIM for high dropout rates $p$.
However, even with the highest considered dropout rate $p=0.75$ the SSIM is increased by $0.52/0.40$ compared to the baseline IG attack for MNIST/CIFAR-10, respectively.

The increase in reconstruction quality for the ViT is even more remarkable.
For dropout rates $p > 0$, the IG based reconstructions yield a SSIM $\approx 0$.
The well-informed attacker WIIG achieves almost perfect reconstructions, \ie SSIM $\approx 1$, for both datasets.
Interestingly, the SSIM increases compared to IG with $p=0$.
This indicates that dropout could, in principle, allow even better reconstructions.
We attribute this effect to the attacker's additional knowledge about $\Psi_C$.
Because the ground truth masks $\Psi_C$ are applied during forward propagation, dropout related zero values in the client and dummy gradients match by default.
The additional information facilitates the problem, as the overall number of gradient values that need to be matched to find an optimal solution is decreased.

\section{DIA -- Dropout Inversion Attack}
\label{sec:dia}
In a realistic scenario the attacker does not have information on the client's dropout masks $\Psi_C$ used during training.
However, we argue that if the attacker finds a close enough approximation $F_{\Psi_A} \approx F_{\Psi_C}$, she still bypasses the privacy inducing effect of dropout.

Assuming an honest-but-curious threat model, the attacker has knowledge of the model architecture and the positions of dropout layers in the model.
To find a realization $F_{\Psi_A}\in \mathcal{F}_p$ that approximates $F_{\Psi_C}$, the attacker has to find dropout masks $\psi_A^{(1)},\dots, \psi_A^{(l)}$ such that $\psi_A^{(i)} \approx \psi_C^{(i)}$ $\forall i = 1, ..., l$, where $\psi_C^{(1)},\dots, \psi_C^{(l)}$ are the dropout masks that were applied during the forward propagation of a local client training step.

\begin{algorithm*}[tb]
	\caption{Dropout Inversion Attack} 
	\label{alg:dia}
	\textbf{Input}: $F$: neural network; $\mathcal{L}$: training loss function; $D$: gradient distance function;\\
	\hspace*{26pt} $\nabla_C=\nabla \mathcal{L}_\theta(F_{\Psi_C}(x),y)$: shared client gradient; $p$: dropout rate; $\eta$: learning rate \\
    \textbf{Output}: $(x', y')$: training data reconstructions; $\Psi_A = \lbrace\psi_A^{(1)},\dots\psi_A^{(l)}\rbrace$: learned dropout masks
	\begin{algorithmic}[1]
	    %\State $x' \gets \mathcal{N}(0,1)$; $y' \gets \mathcal{N}(0,1)$; $M^A_1, ...M^A_l \gets \text{Bernoulli}(p)$;
	    \State $x', y' \gets \mathcal{N}(0,1)$; $\psi_A^{(1)},\dots,\psi_A^{(l)} \gets \text{Bernoulli}(p)$; \Comment{{\color{gray} initialize dummy data and dropout masks}}
 		\While{not converged} \Comment{{\color{gray}reiterate until some optimization criterion is reached}}
 		    \State $\nabla_A \gets \nabla \mathcal{L}_\theta(F_{\Psi_A}(x'),y')$; \Comment{{\color{gray}calculate dummy gradient}}
 		    \State $\mathcal{L}_A \gets D(\nabla_C, \nabla_A)$; \Comment{{\color{gray}calculate gradient distance}}
 		    \State $x' \gets x' - \eta \frac{\delta \mathcal{L}_A}{\delta x'}$; $y' \gets y' - \eta \frac{\delta \mathcal{L}_A}{\delta y'}$; $\psi_A^{(i)} \gets \psi_A^{(i)} - \eta \frac{\delta \mathcal{L}_A}{\delta \psi_A^{(i)}} \text{ } \forall i\in 1,...,l $; \Comment{{\color{gray}update dummy data and dropout masks}}
            % \State $M^A_i \gets \text{clip}(M^A_i,\lbrack 0,1\rbrack)$ %\\Comment{clip masks}
 		\EndWhile
 	\State \textbf{return} $(x', y')$, $\Psi_A$
\end{algorithmic}
\end{algorithm*}

To find a realization $F_{\Psi_A} \approx F_{\Psi_C}$, we propose to optimize the dropout masks $\Psi_A$ used for the forward propagation of dummy data during the gradient inversion attack.
For each dropout layer the corresponding mask $\psi_A^{(i)}$ is initialized randomly from a Bernoulli distribution\footnote{Other initializations are discussed in the technical appendix.} with probability $p$.
Instead of optimizing solely for the dummy data $(x', y')$, the attacker optimizes the dropout masks $\Psi_A$ and the dummy data jointly.
We rewrite the optimization problem as follows:
\begin{equation}
    \label{eq:DIA}
    \small
    \argmin_{ \left( x',y', \Psi_A \right) } D(\nabla \mathcal{L}_\theta(F_{\Psi_C}(x), y),  \nabla \mathcal{L}_\theta(F_{\Psi_A}(x'),y')) + \lambda \Omega.
\end{equation}
The pseudo code for our proposed Dropout Inversion Attack is given as Algorithm~\ref{alg:dia}.

To calculate the dummy gradient $\nabla \mathcal{L}_\theta(F_{\Psi_A}(x'),y'))$ the attacker forwards the dummy image $x'$ through the model realization $F_{\Psi_A}$.
The reconstruction loss between the shared client gradient and dummy gradient is computed and backpropagated.
The gradients for the dummy data $(x', y')$ and the masks $\psi_A^{(i)}$ are calculated and used for optimization. 
Note that elements of the client dropout masks $\psi_C^{(i)} \in \lbrace  0,1 \rbrace$ are binary, whereas the optimized masks $\psi_A^{(i)} \in \lbrack  0,1\rbrack$ are \textit{fuzzy}, since they are adjusted iteratively.
We found that discretization of the masks destabilizes the attack optimization. 
To avoid scaling effects, we clip the masks between $0$ and $1$.
We provide a PyTorch implementation of DIA\footnote{\url{https://github.com/dAI-SY-Group/DropoutInversionAttack}}.

\section{Experiments}
\label{sec:exp}
We use MNIST~\cite{Deng2012} and CIFAR-10~\cite{Krizhevsky2009} datasets that are separated into train and test splits according to the benchmark protocols.
For the attacks we randomly sample a victim client dataset of $128$ images from the training data of one federated client as used in the training.
For experiments on ImageNet~\cite{Russakovsky2015}, we randomly sample $128$ images from different classes from the training dataset.
Client gradients are computed by performing a single training step on victim client data.

Initial experiments are carried out on a Multi Layer Perceptron (MLP)~\cite{Rumelhart1985} and a small version of a Vision Transformer (ViT)~\cite{Dosovitskiy2020}.
For experiments conducted on CNN based architectures we modify the LeNet implementation from~\cite{Zhao2020} and a ResNet-18~\cite{He2016} by adding a dropout layer right before the final fully connected classification layer.
We use IG~\cite{Geiping2020} as baseline attack.
More details on the model architectures, attack configuration and hyperparameter selection can be found in the technical appendix.

To measure reconstruction quality we calculate the \textit{Structural Similarity} (SSIM)~\cite{Wang2004} between the original and reconstructed images.
Higher SSIM indicates higher reconstruction quality.
Additional metrics,~\ie MSE, PSNR and LPIPS, are reported in the technical appendix.

To measure the similarity between the approximated model $F_{\Psi_A}$ and the client model $F_{\Psi_C}$, we compute the \textit{Mean Mask Distance} (MMD) between the optimized dropout masks $\Psi_A$ and the client's dropout masks $\Psi_C$:
\begin{equation}
    \label{eq:mmd}
    \text{MMD}(\Psi_A, \Psi_C) = \frac{1}{l} \sum_{i=1}^{l} ||\psi_A^{(i)}-\psi_C^{(i)}||^2.
\end{equation}
Hence, MMD $=0$ indicates $F_{\Psi_A} = F_{\Psi_C}$, i.e. the attacker model equals the client model.
For each metric we report the average across the $128$ samples of each victim client dataset.

\subsection{Dropout Inversion Attack}
\label{sec:exp_dia}
\begin{figure}
    \centering
    \includegraphics[width=0.94\linewidth]{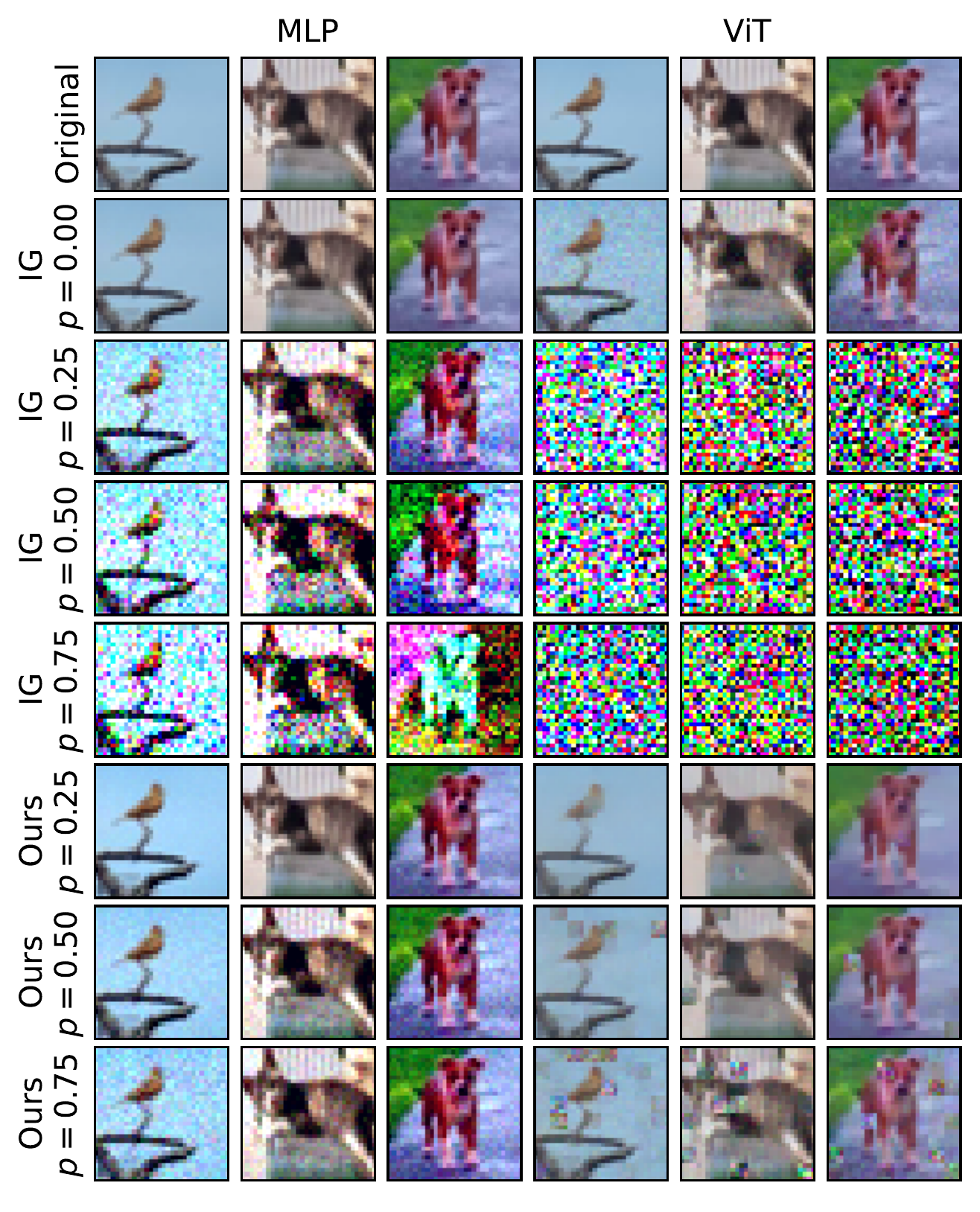}
    \caption{
    \textbf{Example reconstructions} for batchsize $\mathcal{B}=1$ for MLP and ViT on CIFAR-10.
    }
    \label{fig:mlpvit_rec}
\end{figure}

In the first set of experiments the MLP and ViT with batchsizes $\mathcal{B}\in \lbrace 1, 4, 8, 16\rbrace$ are attacked.
Although model utility did not benefit from dropout rates $p > 0.25$ (cf. Tab. \ref{tab:IGAPM}), we choose $p \in \lbrace 0.25, 0.50, 0.75\rbrace$ to assess the efficacy of DIA at increased difficulty.
Example reconstructions are visualized in Fig.~\ref{fig:mlpvit_rec}.
Numeric results are reported in Fig.~\ref{fig:DIAheatmaps}.

We find that, in contrast to IG (cf. Tab.~\ref{tab:IGAPM}), DIA is able to successfully reconstruct client data from shared gradients even if dropout was used during model training.
However, SSIM decreases with increasing dropout rates and batchsizes.
For the MLP with dropout rate $p=0.75$ and batchsize $\mathcal{B} = 16$, DIA based reconstructions achieve a SSIM of $0.8$/$0.63$ on MNIST/CIFAR-10.
For the ViT, increased $p$ and $\mathcal{B}$ affect the reconstruction quality more notably.
SSIM drops below a critical value of $0.6$ if $p \geq 0.5$ and $\mathcal{B} \geq 4$.
However, dropout rates $p\geq 0.25$ also have negative impact on model utility (cf. Tab.~\ref{tab:IGAPM}) and should be avoided for ViTs.

We observe that the joint optimization of dummy data and dropout masks in DIA finds a suitable approximation $F_{\Psi_A} \approx F_{\Psi_C}$ that allows to reconstruct the client data.
For the ViT, DIA based reconstructions achieve smaller SSIM compared to WIIG based reconstructions (cf. Tab.~\ref{tab:IGAPM}),~\ie if the attacker is informed about $\Psi_C$.
In fact, we observe an inverse correlation between SSIM and MMD (cf. Fig.~\ref{fig:DIAheatmaps}(b)), \ie high reconstruction quality (high SSIM) correlates with small mask distance (low MMD), which is a measure for the similarity between $F_{\Psi_A}$ and $F_{\Psi_C}$.
Since dropout masks have to be approximated per sample, increased batchsizes $\mathcal{B}$ increase the number of attack parameters.
In addition, different samples in a batch cause overlapping neuron activations~\cite{Pan2020} and lead to joint gradients.
This increases the difficulty of the attack, as can be observed by decreased SSIM and increased MMD in Fig.~\ref{fig:DIAheatmaps}.

\subsection{Improving Dropout Mask Approximations}
We observe that masks optimized by DIA deviate from client masks with increasing dropout rate and batchsize.
To mitigate this effect, we propose to regularize the optimized masks $\Psi_A$ by $\Omega \left( \Psi_A \right)$ to match the client's dropout rates:
\begin{equation}
    \label{eq:drreg}
    \Omega \left( \Psi_A \right) = \sum_{i=1}^{l} \left| p - \left(1-\frac{||\psi_A^{(i)}||}{n_i}\right)\right|,
\end{equation}
where $n_i$ is the size of dropout mask $\psi_A^{(i)}$.
The client's dropout rate $p$ is part of the model architecture and hence known by the attacker by default (cf. Sec.~\ref{sec:sotF_gia}).

We evaluate the efficacy of $\Omega \left( \Psi_A \right)$ for a fixed dropout rate of $p=0.25$ since higher rates did not improve model utility (cf. Tab.~\ref{tab:IGAPM}).
In addition, we tune the impact of $\Omega \left( \Psi_A \right)$ by weighting with $\lambda_{\text{mask}} \in \lbrace 10^{-4}, 10^{-3}, 10^{-2} \rbrace$.

\begin{figure}[t!]
     \centering
     \begin{minipage}[c]{\linewidth}
     \begin{subfigure}{\linewidth}
         \centering
         \includegraphics[width=\linewidth]{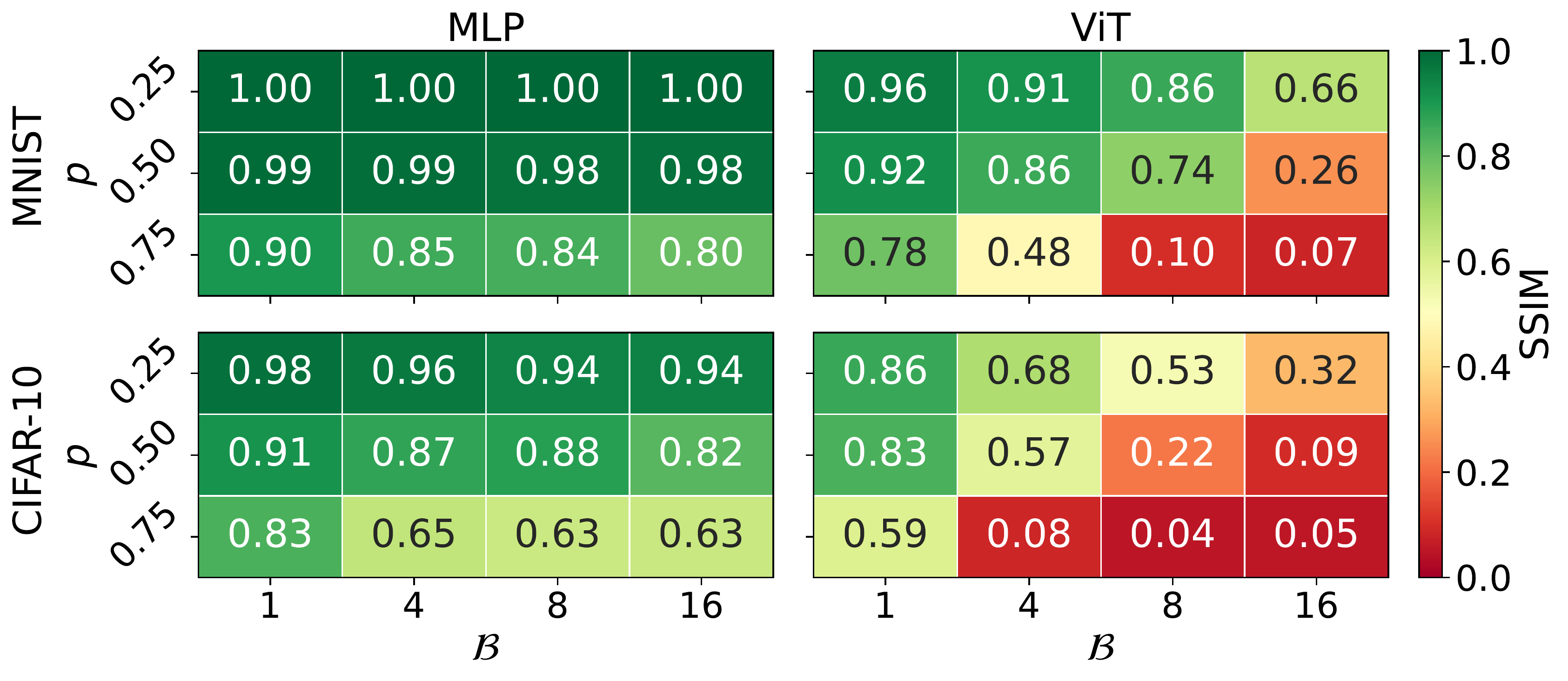}  
         \caption{SSIM for different dropout rates $p$ and batchsizes $\mathcal{B}$.}
     \end{subfigure}
     \end{minipage}
     \begin{minipage}[c]{\linewidth}
     \begin{subfigure}{\linewidth}
         \centering
         \vspace{10pt}
         \includegraphics[width=\linewidth]{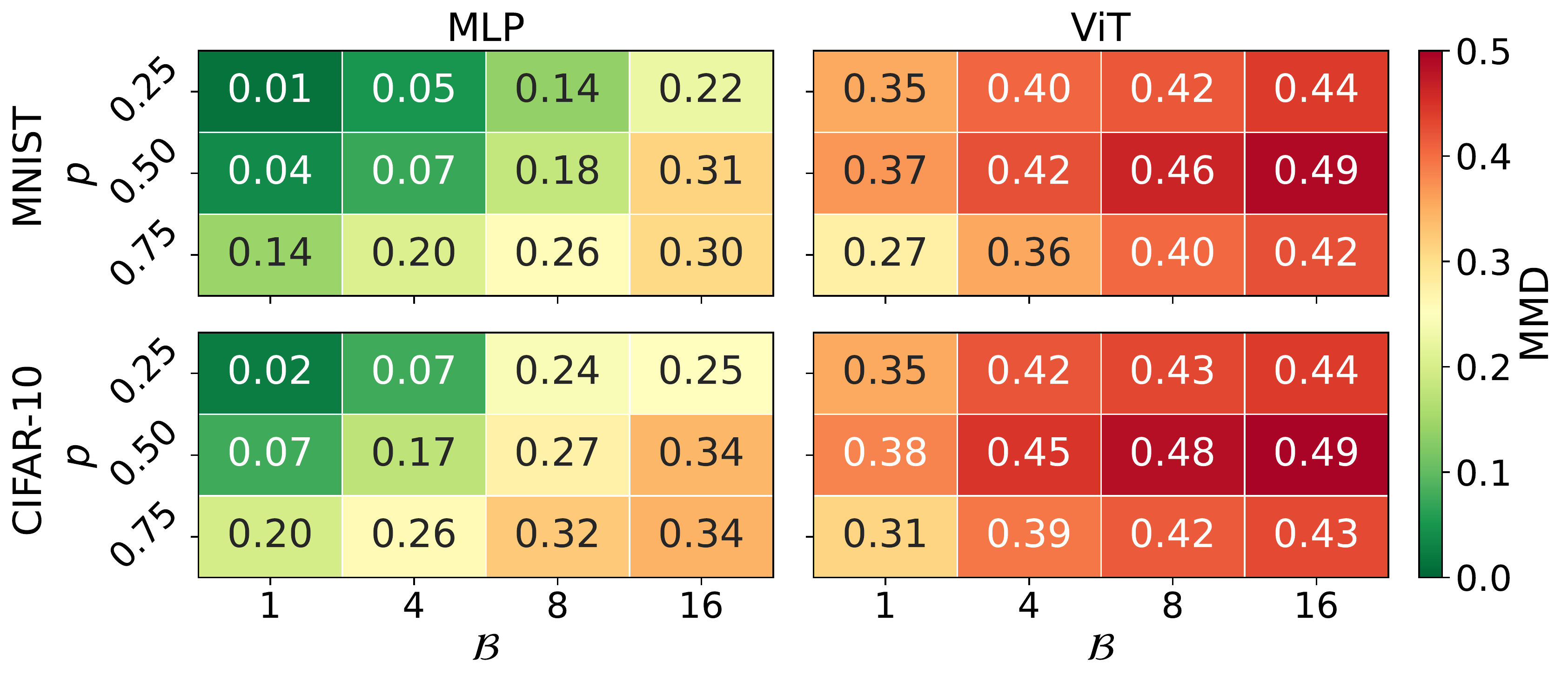}  
         \caption{MMD for different dropout rates $p$ and batchsizes $\mathcal{B}$.}
     \end{subfigure}
     \end{minipage}
     \begin{minipage}[c]{\linewidth}
     \begin{subfigure}{\linewidth}
         \centering
         \vspace{10pt}
         \includegraphics[width=\linewidth]{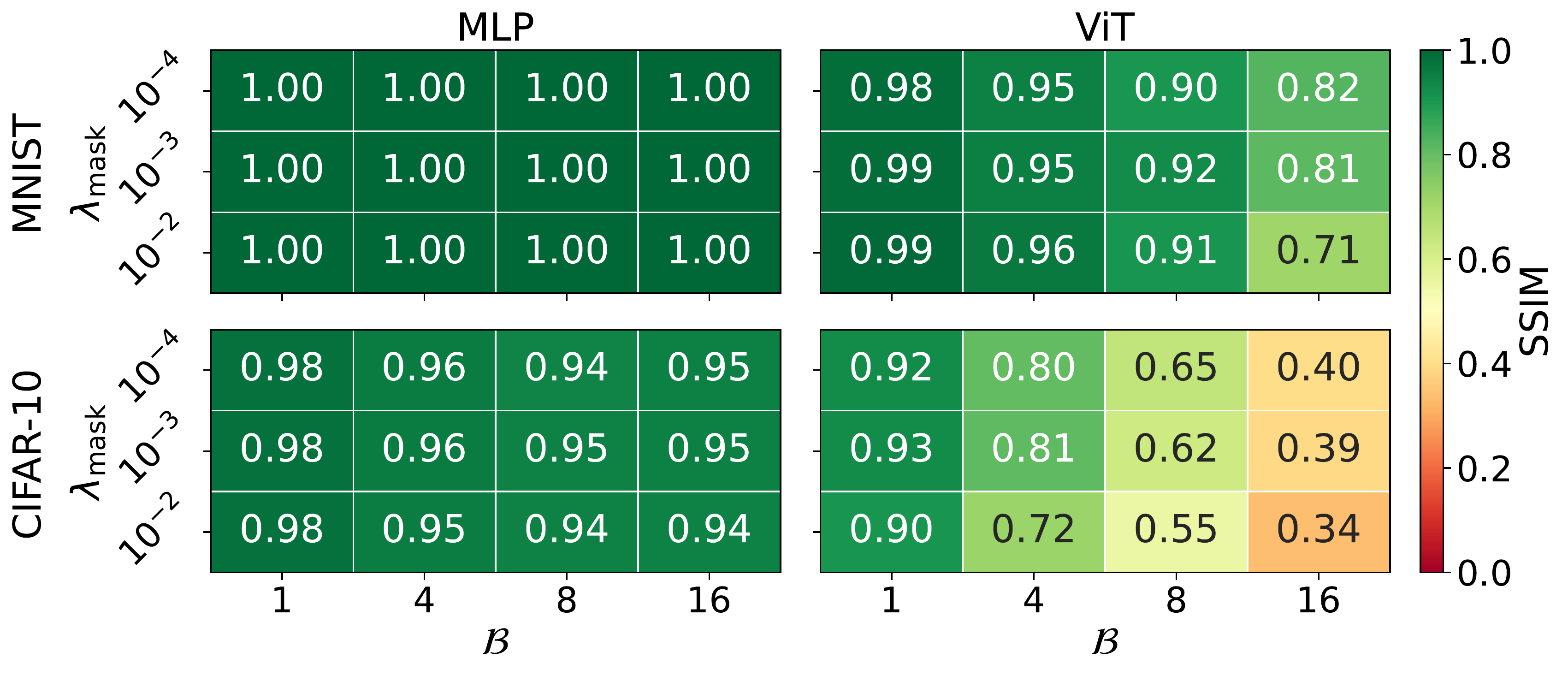}  
         \caption{SSIM for different regularization parameter selections $\lambda_{\text{mask}}$ and batchsizes $\mathcal{B}$ with fixed dropout rate $p=0.25$.}
     \end{subfigure}
     \end{minipage}
     \caption{
     \textbf{DIA reconstruction results} for MLP (left) and ViT (right) on MNIST and CIFAR-10.}
     \label{fig:DIAheatmaps}
\end{figure}

The results of this mask regularization are displayed in Fig.~\ref{fig:DIAheatmaps} (c).
As the SSIM for the MLP is already close to $1$, only marginal improvement is observed upon addition of $\Omega \left( \Psi_A \right)$.
For the ViT the added mask regularization shows a notable increase in SSIM and hence improved reconstruction quality, especially for $\mathcal{B} > 1$.
Since we find our proposed mask regularization to improve reconstruction quality, we utilize it with $\lambda_\text{mask} = 10^{-4}$ for all further experiments.

\subsection{Attacking Dropout at Higher Scales}
Since DIA jointly optimizes for dummy data and dropout masks, the number of optimized parameters increases with (1) the number of dropout layers $l$ in the model and (2) the input batchsize.
ViTs also use an image patch embedding; hence, both input dimensions and batchsize further influence the number of parameters.
We therefore want to investigate the applicability of our proposed attack on a state of the art sized ViT-B/16 and a practical image classification dataset, \ie ImageNet.
Following the recommendations of the original ViT paper, we apply a dropout rate of $p=0.1$ for the ViT-B/16~\cite{Dosovitskiy2020}.

\begin{table}[!t]
\centering
\resizebox{0.65\linewidth}{!}{%
\begin{tabular}{c|c|c|c|c}
\toprule
 &  & & IG & Ours \\
 & Model & $p$ & SSIM $\uparrow$ & SSIM $\uparrow$  \\
\midrule
\multirow[c]{2}{*}{\rotatebox{90}{IN}} & \multirow[c]{2}{*}{ViT-B/16} & 0.00& {\cellcolor[HTML]{E8F59F}} \color[HTML]{000000} \bfseries 0.56& {\cellcolor[HTML]{FFFFFF}} \color[HTML]{000000} - \\
 &  & 0.10& {\cellcolor[HTML]{A90426}} \color[HTML]{F1F1F1} \itshape 0.01& {\cellcolor[HTML]{98D368}} \color[HTML]{000000} \itshape \bfseries 0.72\\
\midrule
\multirow[c]{8}{*}{\rotatebox{90}{MNIST}} & \multirow[c]{4}{*}{LeNet} & 0.00& {\cellcolor[HTML]{0C7F43}} \color[HTML]{F1F1F1} \bfseries 0.95& {\cellcolor[HTML]{FFFFFF}} \color[HTML]{000000} - \\
 &  & 0.25& {\cellcolor[HTML]{E5F49B}} \color[HTML]{000000} 0.57& {\cellcolor[HTML]{0F8446}} \color[HTML]{F1F1F1} 0.94\\
 &  & 0.50& {\cellcolor[HTML]{FEE08B}} \color[HTML]{000000} 0.40& {\cellcolor[HTML]{0C7F43}} \color[HTML]{F1F1F1} \bfseries 0.95\\
 &  & 0.75& {\cellcolor[HTML]{F67F4B}} \color[HTML]{F1F1F1} 0.23& {\cellcolor[HTML]{0F8446}} \color[HTML]{F1F1F1} 0.94\\
\cline{2-5}
 & \multirow[c]{4}{*}{ResNet} & 0.00& {\cellcolor[HTML]{279F53}} \color[HTML]{F1F1F1} 0.88& {\cellcolor[HTML]{FFFFFF}} \color[HTML]{000000} - \\
 &  & 0.25& {\cellcolor[HTML]{FED07E}} \color[HTML]{000000} 0.37& {\cellcolor[HTML]{0F8446}} \color[HTML]{F1F1F1} 0.94\\
 &  & 0.50& {\cellcolor[HTML]{EE613E}} \color[HTML]{F1F1F1} 0.18& {\cellcolor[HTML]{118848}} \color[HTML]{F1F1F1} \itshape 0.93\\
 &  & 0.75& {\cellcolor[HTML]{D22B27}} \color[HTML]{F1F1F1} \itshape 0.09& {\cellcolor[HTML]{118848}} \color[HTML]{F1F1F1} \itshape 0.93\\
\midrule
\multirow[c]{8}{*}{\rotatebox{90}{CIFAR-10}} & \multirow[c]{4}{*}{LeNet} & 0.00& {\cellcolor[HTML]{219C52}} \color[HTML]{F1F1F1} \bfseries 0.89& {\cellcolor[HTML]{FFFFFF}} \color[HTML]{000000} - \\
 &  & 0.25& {\cellcolor[HTML]{FFF8B4}} \color[HTML]{000000} 0.48& {\cellcolor[HTML]{219C52}} \color[HTML]{F1F1F1} \bfseries 0.89\\
 &  & 0.50& {\cellcolor[HTML]{FDB768}} \color[HTML]{000000} 0.32& {\cellcolor[HTML]{279F53}} \color[HTML]{F1F1F1} 0.88\\
 &  & 0.75& {\cellcolor[HTML]{F57245}} \color[HTML]{F1F1F1} 0.21& {\cellcolor[HTML]{279F53}} \color[HTML]{F1F1F1} 0.88\\
\cline{2-5}
 & \multirow[c]{4}{*}{ResNet} & 0.00& {\cellcolor[HTML]{C5E67E}} \color[HTML]{000000} 0.64& {\cellcolor[HTML]{FFFFFF}} \color[HTML]{000000} - \\
 &  & 0.25& {\cellcolor[HTML]{FBA05B}} \color[HTML]{000000} 0.28& {\cellcolor[HTML]{A0D669}} \color[HTML]{000000} 0.71\\
 &  & 0.50& {\cellcolor[HTML]{E54E35}} \color[HTML]{F1F1F1} 0.15& {\cellcolor[HTML]{A5D86A}} \color[HTML]{000000} \itshape 0.70\\
 &  & 0.75& {\cellcolor[HTML]{CC2627}} \color[HTML]{F1F1F1} \itshape 0.08& {\cellcolor[HTML]{A0D669}} \color[HTML]{000000} 0.71\\
\bottomrule
\end{tabular}
 }
\caption{
SSIM computed from gradients with $\mathcal{B}=1$ attacked with IG and DIA (Ours) for ViT-B/16 on ImageNet (IN) as well as LeNet and ResNet on MNIST and CIFAR-10.
Arrows indicate direction of improvement.
Bold and italic formatting highlight best and worst results respectively.
}
\label{tab:OtherModels}
\end{table}

\begin{figure}[!t]
    \centering
    \includegraphics[width=0.94\linewidth]{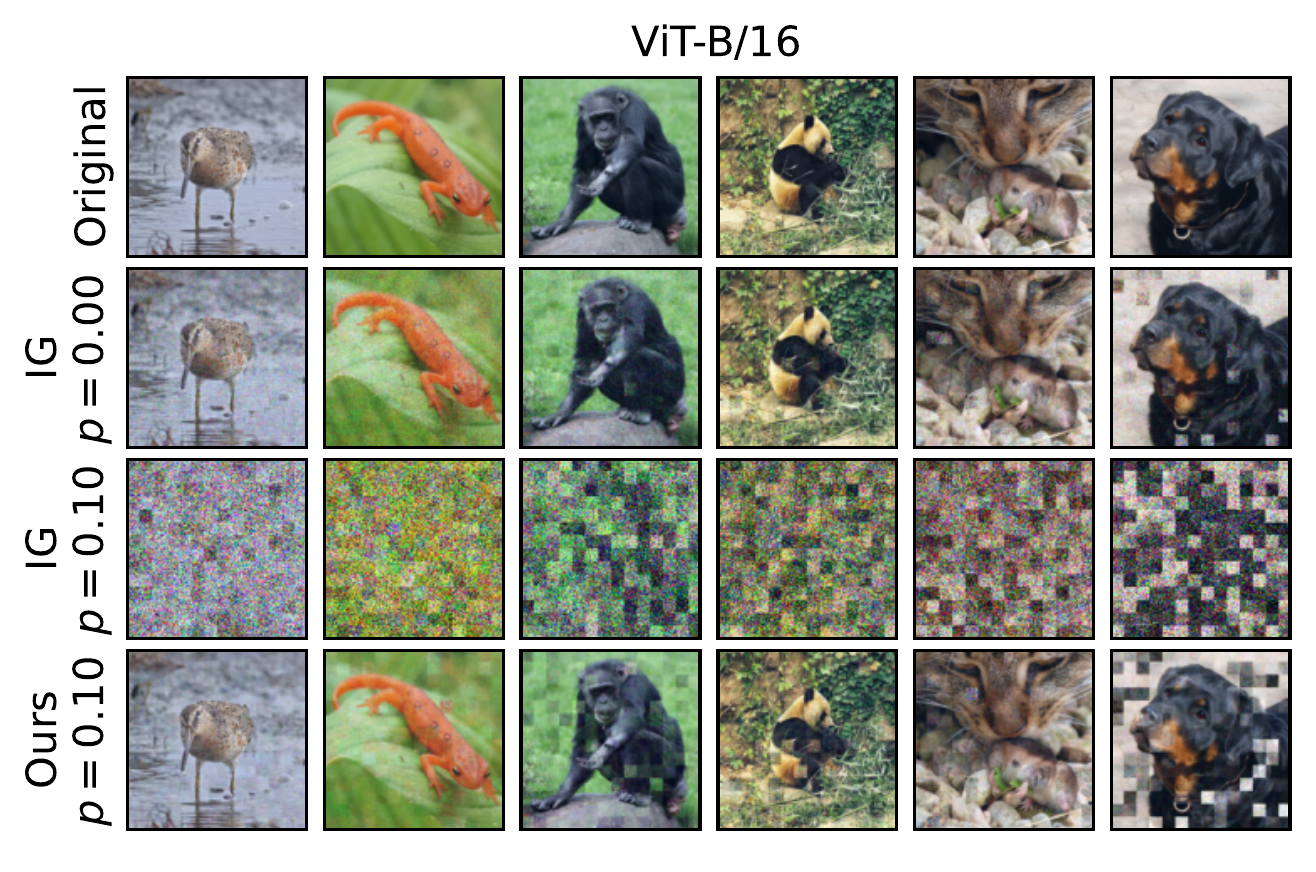}
    \caption{
    \textbf{Example reconstructions} for batchsize $\mathcal{B}=1$ for ViT-B/16 on ImageNet.
    }
    \label{fig:vitb16}
\end{figure}

Tab.~\ref{tab:OtherModels} shows that even for such a low dropout rate $p$, IG is not able to reconstruct the data.
In comparison, DIA based reconstructions achieve a SSIM of $0.72$.
As observed before, the reconstruction quality for DIA with dropout is higher compared to IG without dropout.
Reconstruction examples are visualized in Fig.~\ref{fig:vitb16}.

\subsection{Attacking Dropout in CNNs}
\begin{figure}[!t]
    \centering
    \includegraphics[width=0.94\linewidth]{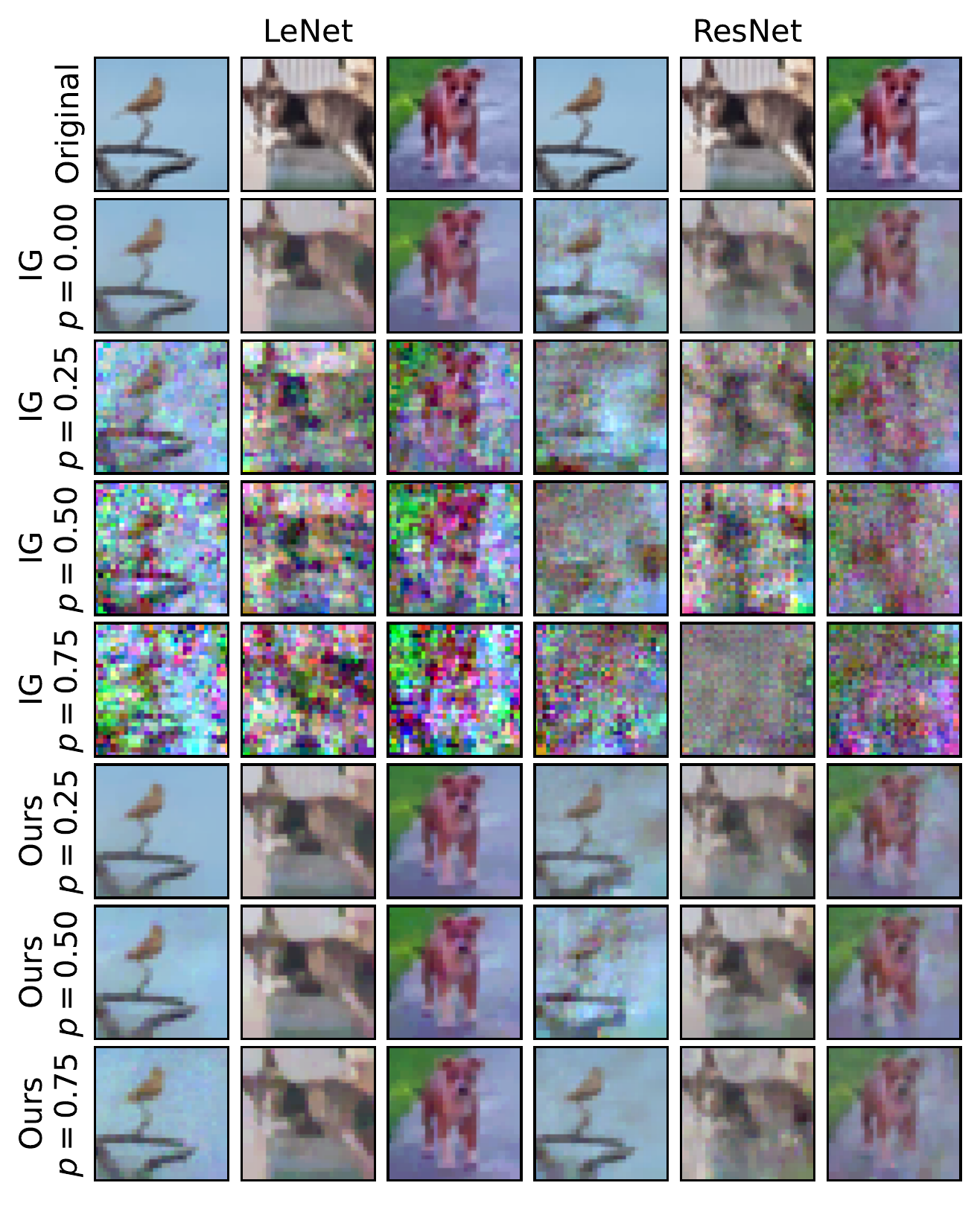}
    \caption{
    \textbf{Example reconstructions} for batchsize $\mathcal{B}=1$ for LeNet and ResNet on CIFAR-10.
    }
    \label{fig:cnns}
\end{figure}

Recent work commonly evaluates gradient inversion attacks for CNN based architectures like LeNet and ResNet~\cite{Zhao2020, Geiping2020, Wei2020, Yin2021}.
Furthermore, a drop in reconstruction quality was reported when dropout is used before the output layer of a LeNet~\cite{Zheng2021}.
We therefore investigate the efficacy of our proposed attack on these CNN based classifiers if dropout is applied before the output layer.
The results in Tab.~\ref{tab:OtherModels} confirm that for the baseline IG attack reconstruction quality decreases for increased dropout rates for both model architectures.
In contrast, when DIA is used as attack, client data is successfully reconstructed regardless of enabled dropout.
Moreover, compared to the MLP and ViT architectures, SSIM remains at the same level even with increased dropout rates.
We argue that since the CNN based architectures utilize only one dropout layer, the gradients of the other layers retain sufficient information for reconstruction.
Reconstruction examples are visualized in Fig.~\ref{fig:cnns}. 

\section{Conclusion}
\label{sec:concl}
Recent work suggests that dropout in neural networks improves data privacy during federated learning, because it seems to prevent gradient inversion attacks.
We formalize the impact of dropout on such inversion attacks based on specific realizations of a stochastic model.
Dropout causes an inherent mismatch between the model realizations of the attacker and client, which in turn prevents reconstruction of client data.
However, this offers a premature sense of security, because an attacker can still reconstruct client data either by being informed about the client's dropout masks or by approximating them. 
To showcase the vulnerability of dropout protected neural networks, we formulate a novel Dropout Inversion Attack (DIA) that jointly optimizes for client data and dropout masks to approximate the client's model realization.
We conduct an extensive systematic empirical study to investigate the impact of dropout on four seminal model architectures and three image classification datasets of increasing complexity.
We show that our proposed attack successfully bypasses the seemingly induced protection of dropout and allows to reconstruct data with high fidelity.
Although we evaluate our proposed attack solely in an image classification setting, we expect DIA to be universally applicable since the underlying mechanism can be trivially integrated into other iterative inversion attacks.
We confirm that the strategic use of architectural features, such as dropout, cannot be assumed to sufficiently protect client privacy in federated learning scenarios.
We conclude that a combination of complementary defense mechanisms should be applied in order to protect privacy and maintain model utility.

\section{Acknowledgments}
\label{sec:ack}
% Placeholder for final submission.
We are funded by the Thuringian Ministry for Economic Affairs, Science and Digital Society (Grant: 5575/10-3).

\bibliography{bibliography}

\clearpage

\begin{strip}  
\vspace{-30pt}
\begin{center}
      {\Large \bf Supplementary Material}
      \vspace*{12pt}
\end{center}     
\end{strip}    

\appendix
%%%%%%%%% BODY TEXT

\section{Overview}
Section~\ref{sec:code_apx} references to code repositories and datasets we used for our studies.
The technical appendix gives more details on parameter choices for model architectures and experiments.
We also discuss further options for the initialization of dropout masks for our proposed DIA.
Furthermore, we report additional metrics and example reconstructions for all our experiments.

\section{Code and Data}
\label{sec:code_apx}
We base our experiments on the PyTorch implementation of IG\footnote[1]{\url{https://github.com/JonasGeiping/invertinggradients}}~\cite{Geiping2020}.
An implementation of our proposed DIA is available on GitHub\footnote[2]{\url{https://github.com/dAI-SY-Group/DropoutInversionAttack}}.

All our experiments were conducted on publicly available datasets: MNIST~\cite{Deng2012}, CIFAR-10~\cite{Krizhevsky2009} and ImageNet~\cite{Russakovsky2015}.
Victim client datasets for reconstruction are randomly sampled from the corresponding training data splits.
We provide these victim datasets in our public code repository\footnotemark[2].

Experiments were conducted on NVIDIA GeForce RTX 2080 Ti GPU, Intel(R) Xeon(R) Silver 4114 CPU and $64$ GB of working memory under \textit{Linux-5.4.0-91-generic-x86\_64-with-glibc2.31} OS.
We use Python version $3.9.10$ and PyTorch version $1.10.2$.
A more detailed list of names and versions of used libraries and frameworks are also available in our public code repository\footnotemark[2].

\section{Technical Appendix}
\label{sec:tech_apx}
\subsection{Models}
Most of our experiments are carried out on a Multi Layer Perceptron (MLP)~\cite{Rumelhart1985} and a small version of a Vision Transformer (ViT)~\cite{Dosovitskiy2020}.
The MLP consists of $3$ hidden layers with width $512$ that are followed by GeLU activation and a dropout layer.
For our versions of the ViT we use a publicly available implementation\footnote[3]{\url{https://github.com/lucidrains/vit-pytorch}}. 
For architecture parameters we choose an embedding patchsize of $4$, $4$ attention blocks with $16$ attention heads each, a hidden size and MLP width of $256$ whereas each MLP head is again followed by GeLU activation and a dropout layer.
For the ViT-B/16 we follow the architecture descriptions in~\cite{Dosovitskiy2020}.

For experiments conducted on CNN based architectures we modify the LeNet from~\cite{Zhao2020} (implementation available in\footnotemark[1]) and a ResNet-18~\cite{He2016} (PyTorch torchvision implementation) by adding a dropout layer right before the final fully connected classification layer.
All models utilize a final classification layer with $10$ output neurons ($1000$ for ImageNet) and softmax activation.
Moreover, all models are initialized randomly.

\subsection{Experiment Parameters}
We configured our IG based attack experiments as follows: 
\begin{itemize}
    \item Dummy data is initialized from Gaussian distribution.
        We assume label information to be known for each attacked sample as it can be analytically reconstructed from gradients of cross-entropy loss functions \wrt weights of fully connected layers with softmax activation~\cite{Geiping2020, Zhao2020, Wei2020}.
    \item Cosine distance is used as reconstruction loss function $D$.
    \item Total variation is used as regularization term with weight $\lambda_{TV}=10^{-5}$ for MLP and ViT as well as $\lambda_{TV}=10^{-2}$ for LeNet and ResNet.
    \item We use Adam~\cite{Kingma2014} optimizer with initial learning rate $0.1$. We reduce the learning rate by a factor of $0.1$ if the reconstruction loss plateaus for $800$ attack iterations.
    \item To save computational resources, we stop the reconstruction optimization if one of the following termination conditions is reached:
    \begin{itemize}
        \item The reconstruction loss falls below a value of $10^{-5}$.
        \item The reconstruction loss does not decrease for $4'000$ iterations.
        \item A maximum of $20'000$ iterations is reached.
    \end{itemize}
\end{itemize}
If not otherwise stated, for experiments with DIA we use the same base configurations as described above.

For test experiments with iterative APRIL~\cite{Lu2021} on the ViT (cf. section~\ref{sec:add_exp}), we use Euclidean distance for $D$ and add the cosine distance between client and dummy gradients of the positional embedding as regularization term.
We weigh the regularization term with $\lambda_{\mathrm{pos}}=10^{-4}$.

Consistent with related work~\cite{Geiping2020, Enthoven2020a, Yin2021, Kaissis2021, Jin2021, Scheliga2022a, Scheliga2022b, Zhang2022, Gupta2022}, 
we limit our threat model to target gradient leakage for only one local training iteration.
If there are multiple local training iterations where each iteration applies a different set of dropout masks, the attack complexity would massively increase.

\subsection{Additional Metrics and Experimental Results}
\label{sec:add_exp}
Besides \textit{Structural Similarity} (SSIM)~\cite{Wang2004}, we report \textit{Mean Squared Error} (MSE), \textit{Peak Signal-to-Noise Ratio} (PSNR), and a \textit{Learned Perceptual Image Patch Similarity} (LPIPS)~\cite{Zhang2018} to measure reconstruction quality of image reconstructions.
Higher SSIM and PSNR as well as lower MSE and LPIPS indicates higher reconstruction quality.
Furthermore, we report the number of parameters that the attacker is optimizing during attack.
For experiments that use our proposed DIA we also report \textit{Mean Mask Distance} (MMD) and the \textit{Throughput Rate Distance} (TRD) of the learned dropout masks $\Psi_A$ to measure the similarity between the approximated attacker model $F_{\Psi_A}$ and the client model realization $F_{\Psi_C}$.
TRD determines how close the \textit{throughput rate} of the learned fuzzy masks $\Psi_A$ are to the real dropout rate $p$ of the ground truth client masks $\Psi_C$.
\begin{equation}
    \label{eq:trd}
    \text{TRD}(\Psi_A, p) = \frac{1}{l} \sum_{i=1}^{l} \left| p - \left(1-\frac{||\psi_A^{(i)}||}{n_i}\right)\right|, 
\end{equation}
where $n_i$ is the size of dropout mask $\psi_A^{(i)}$.
MMD and TRD values close to $0$ indicate good approximations $F_{\Psi_A} \approx F_{\Psi_C}$.
For each metric we report the average and standard deviation across the $128$ samples of each victim dataset.

Tab.~\ref{tab:mask_init} compares different mask initialization schemes for DIA.
More details can be found in section~\ref{sec:mask_init}.
Tab.~\ref{tab:APRIL} compares IG~\cite{Geiping2020} and the iterative APRIL attack~\cite{Lu2021} as well as the correspondingly extend versions of our DIA attack for the ViT architecture.
Note that these results have been obtained in the course of some preliminary experiments that used a smaller victim dataset with just $10$ samples and batchsize $\mathcal{B} = 1$.
Hence, slight deviations in the results are to be expected.

\begin{figure*}[!t]
    \centering
    \includegraphics[width=\linewidth]{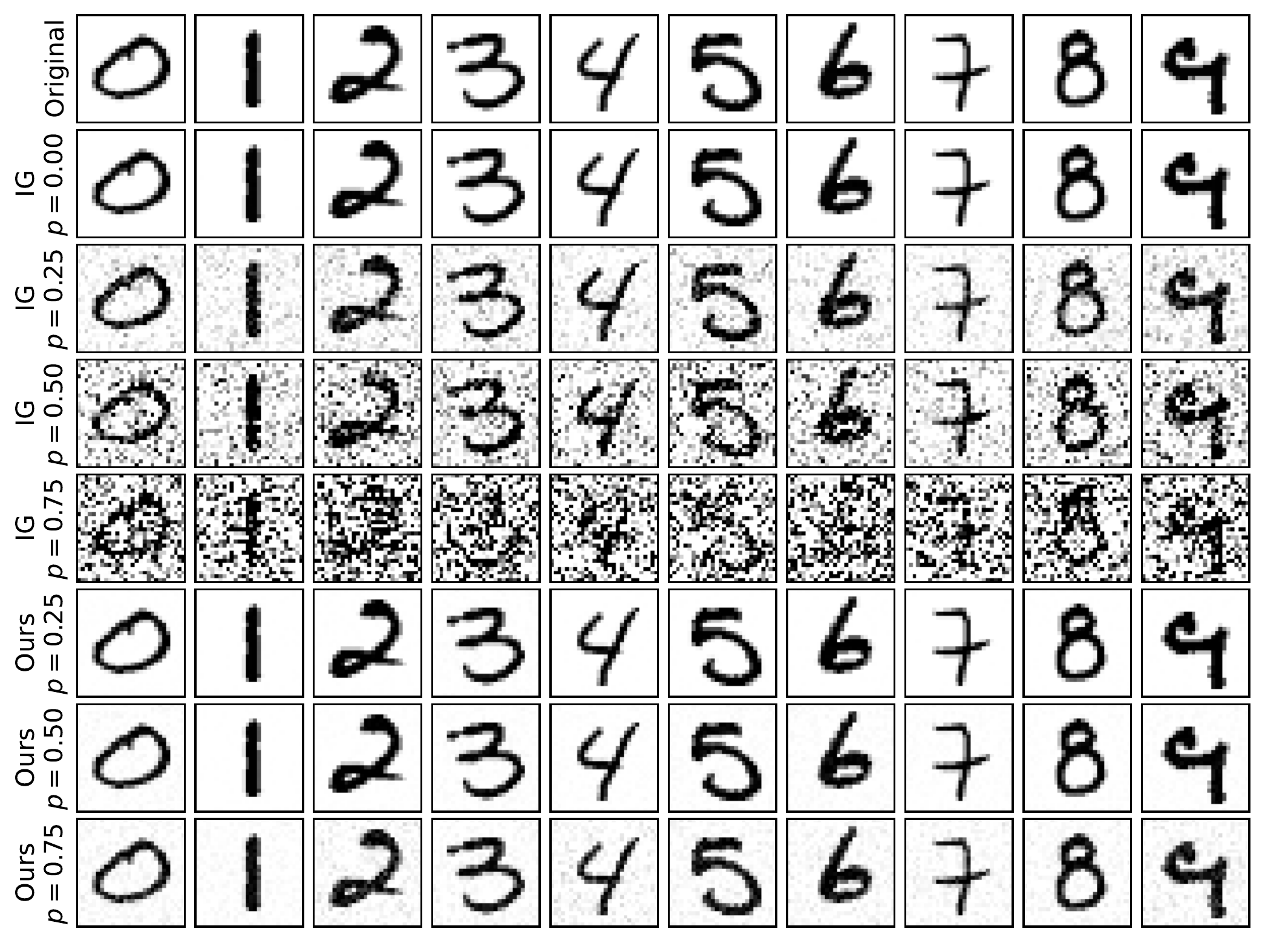}
    \caption{
    Example reconstructions for batchsize $\mathcal{B}=1$ for MLP on MNIST.
    }
    \label{fig:apx_mnist_mlp}
\end{figure*}

\begin{figure*}[!t]
    \centering
    \includegraphics[width=\linewidth]{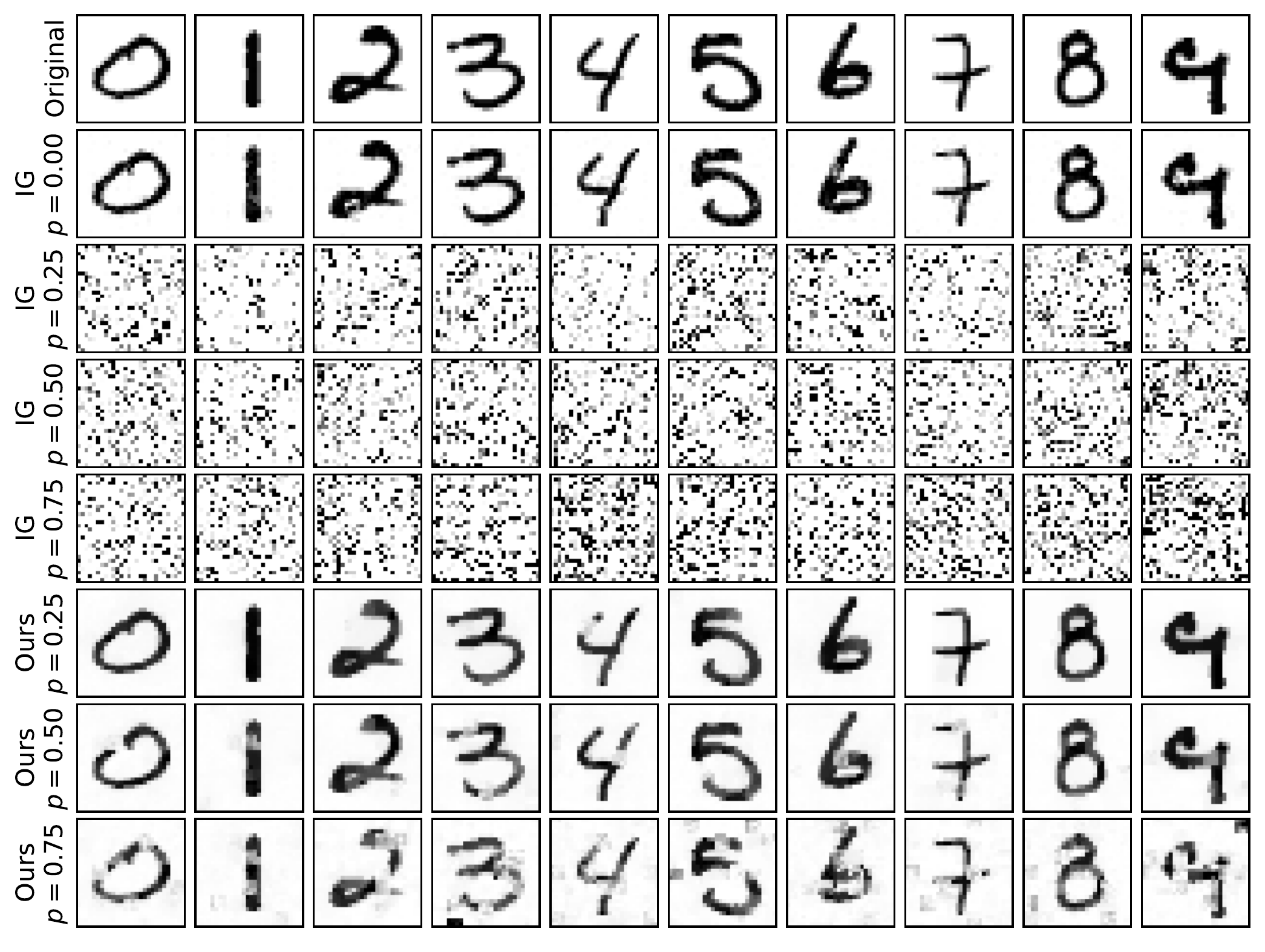}
    \caption{
    Example reconstructions for batchsize $\mathcal{B}=1$ for ViT on MNIST.
    }
    \label{fig:apx_mnist_vit}
\end{figure*}

\begin{figure*}[!t]
    \centering
    \includegraphics[width=\linewidth]{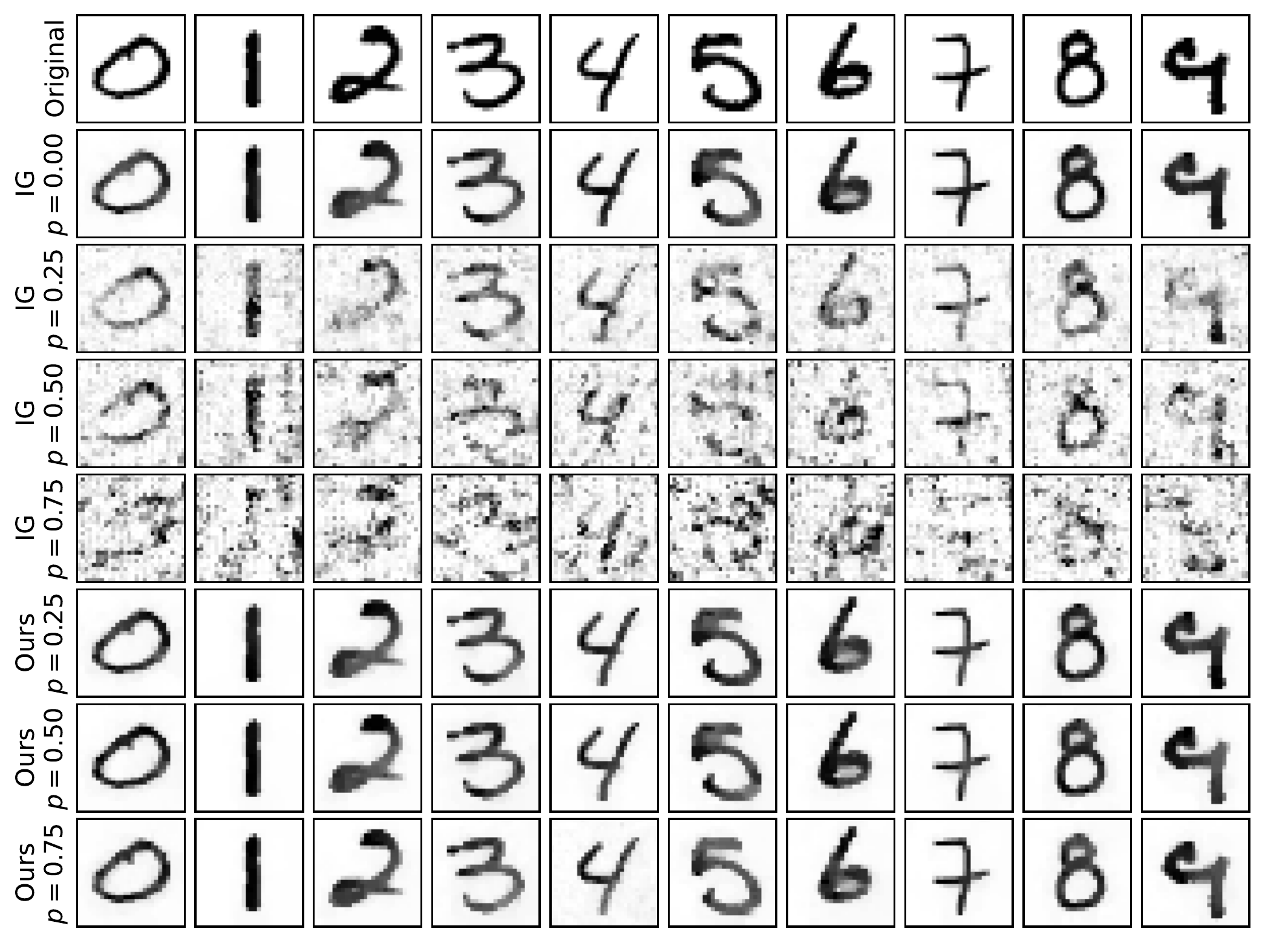}
    \caption{
    Example reconstructions for batchsize $\mathcal{B}=1$ for LeNet on MNIST.
    }
    \label{fig:apx_mnist_le}
\end{figure*}

\begin{figure*}[!t]
    \centering
    \includegraphics[width=\linewidth]{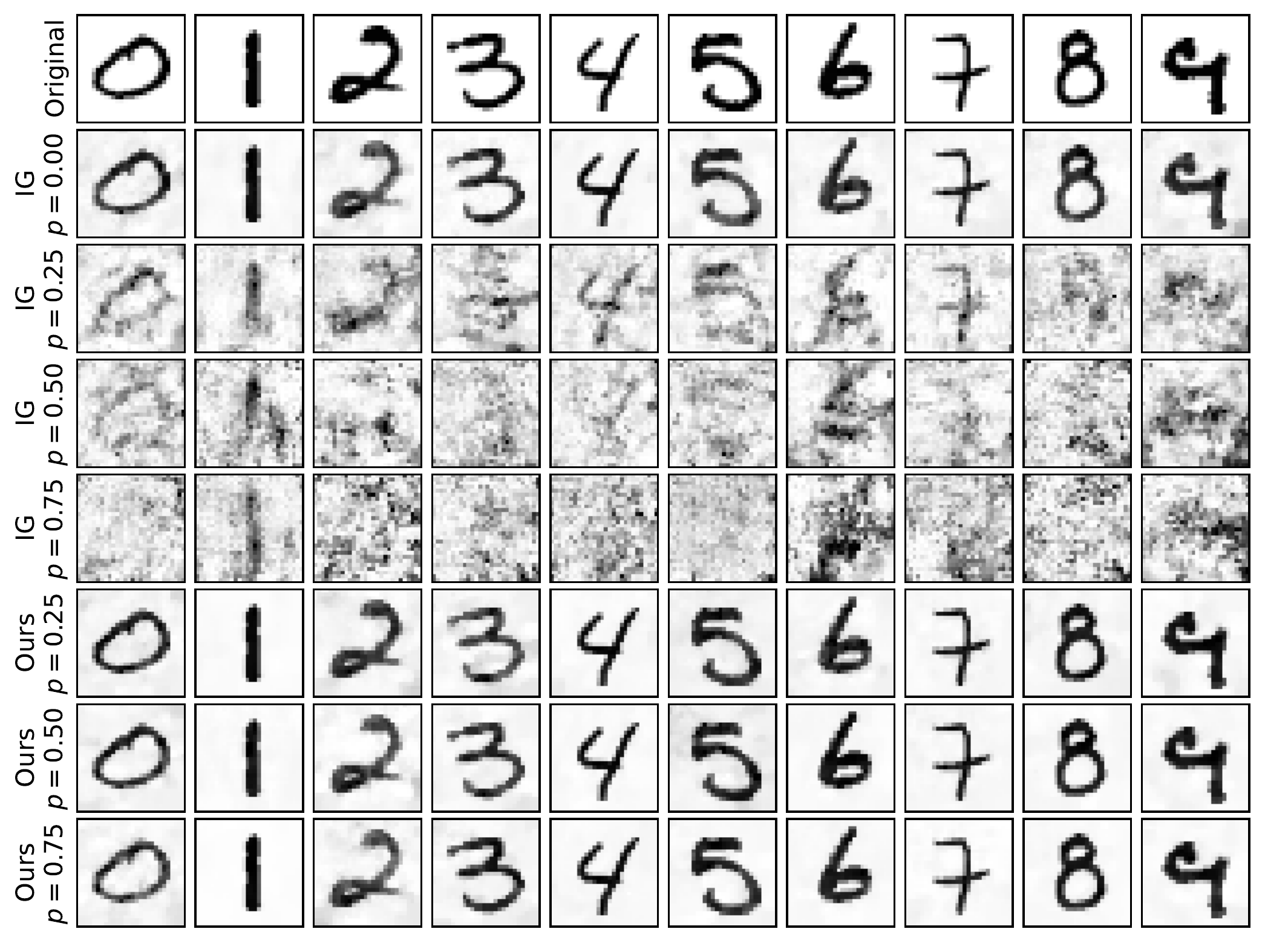}
    \caption{
    Example reconstructions for batchsize $\mathcal{B}=1$ for ResNet on MNIST.
    }
    \label{fig:apx_mnist_res}
\end{figure*}

\begin{figure*}[!t]
    \centering
    \includegraphics[width=\linewidth]{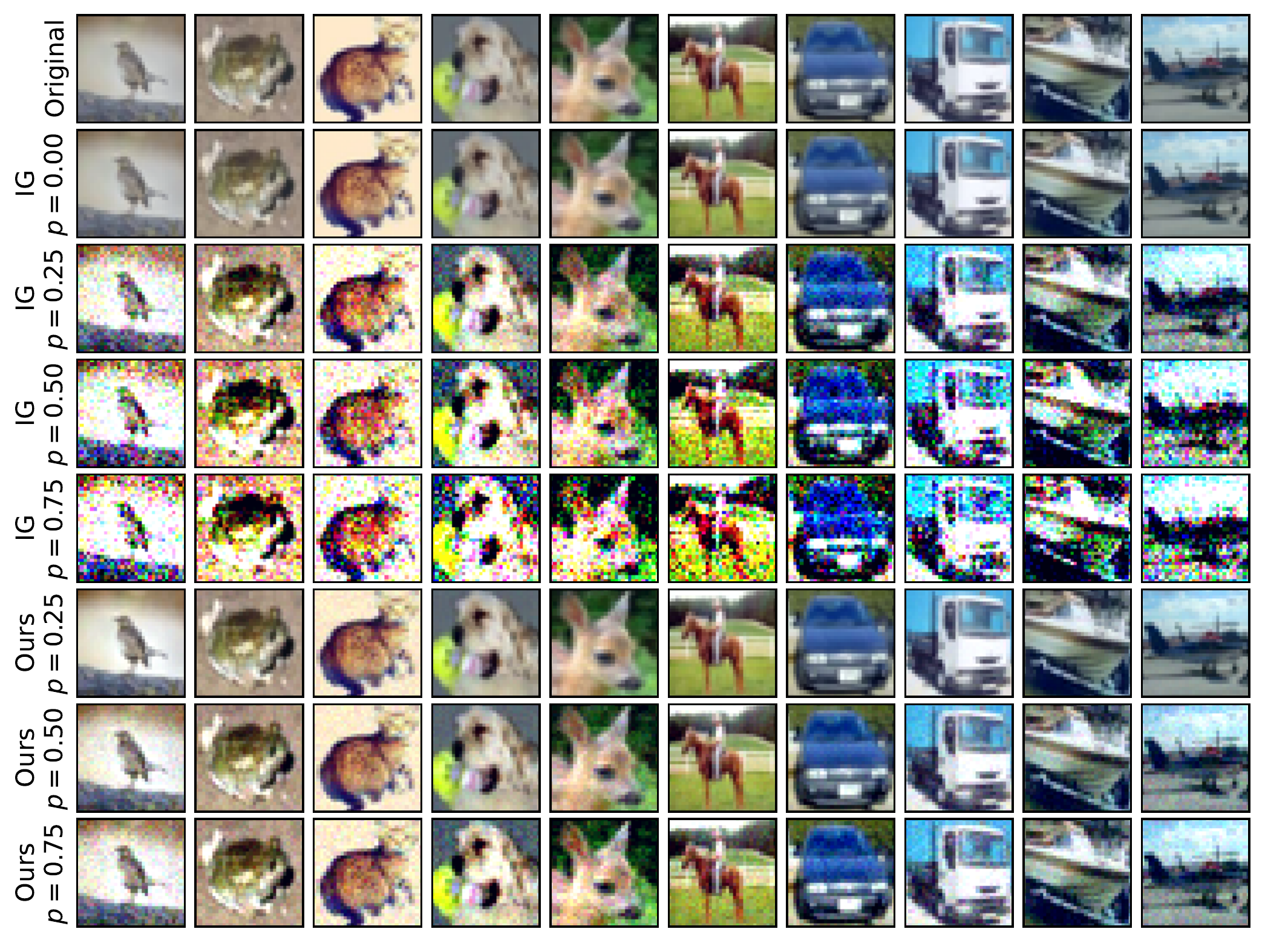}
    \caption{
    Example reconstructions for batchsize $\mathcal{B}=1$ for MLP on CIFAR-10.
    }
    \label{fig:apx_cifar_mlp}
\end{figure*}

\begin{figure*}[!t]
    \centering
    \includegraphics[width=\linewidth]{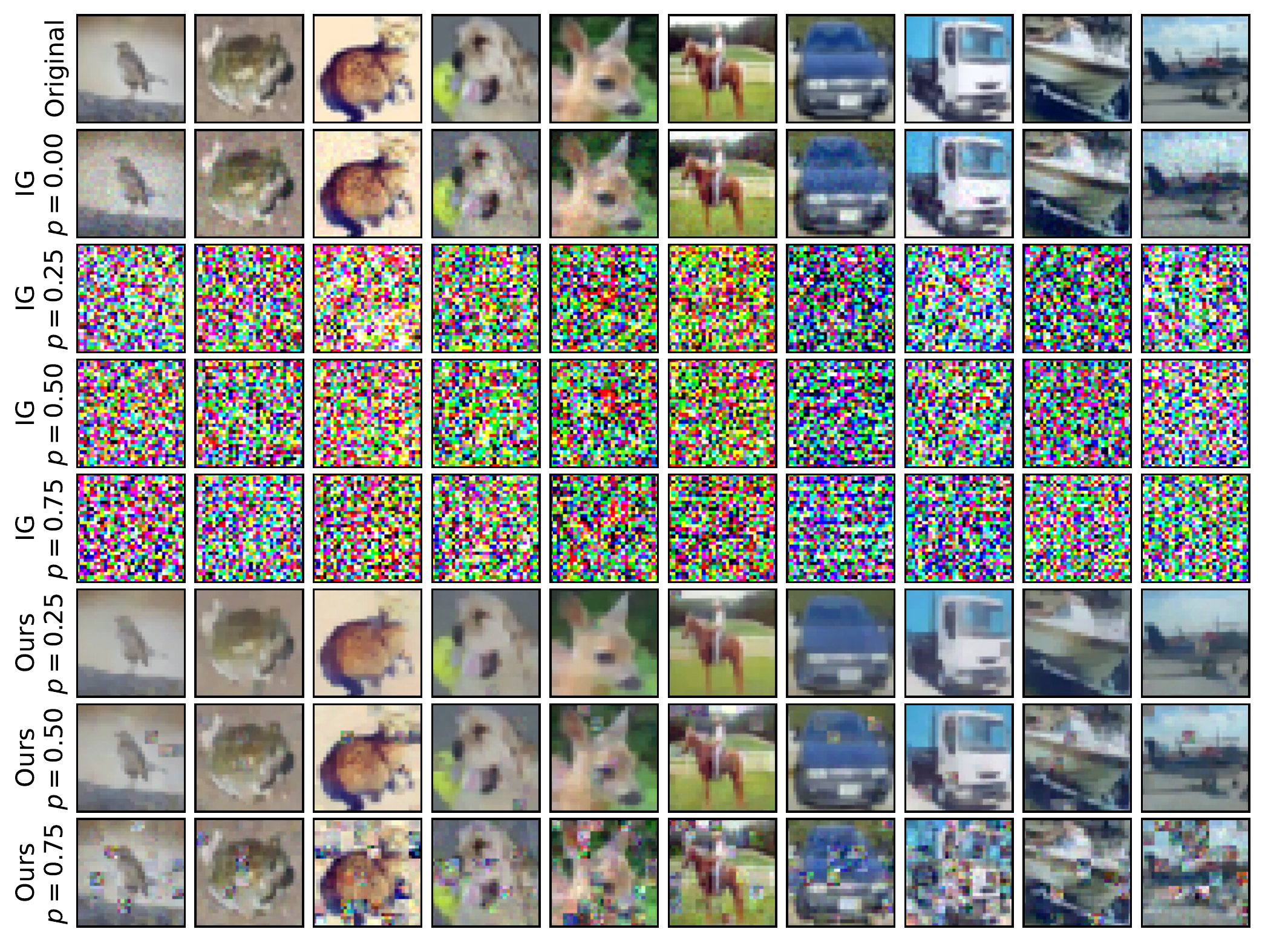}
    \caption{
    Example reconstructions for batchsize $\mathcal{B}=1$ for ViT on CIFAR-10.
    }
    \label{fig:apx_cifar_vit}
\end{figure*}

\begin{figure*}[!t]
    \centering
    \includegraphics[width=\linewidth]{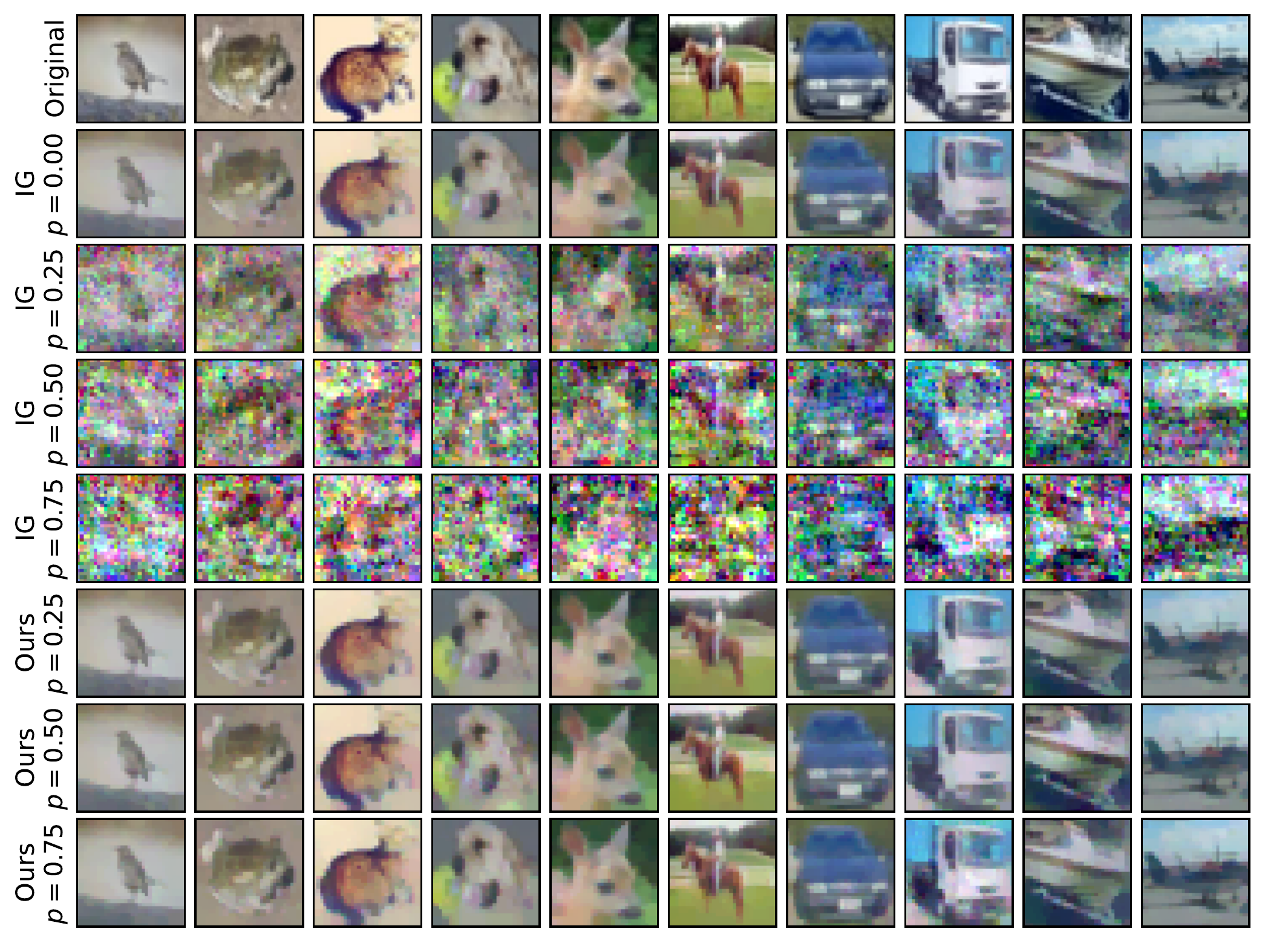}
    \caption{
    Example reconstructions for batchsize $\mathcal{B}=1$ for LeNet on CIFAR-10.
    }
    \label{fig:apx_cifar_le}
\end{figure*}

\begin{figure*}[!t]
    \centering
    \includegraphics[width=\linewidth]{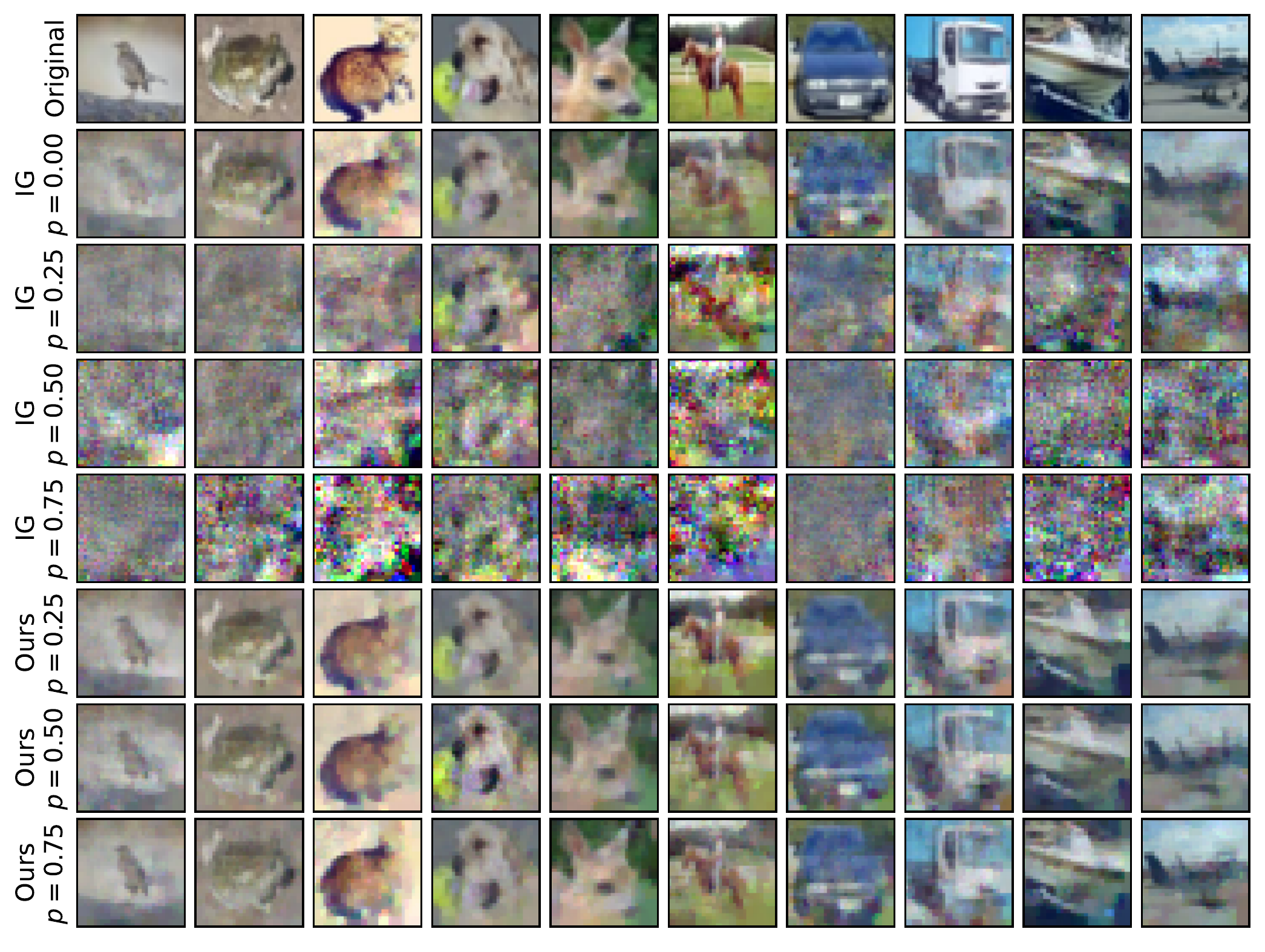}
    \caption{
    Example reconstructions for batchsize $\mathcal{B}=1$ for ResNet on CIFAR-10.
    }
    \label{fig:apx_cifar_res}
\end{figure*}

\begin{figure*}[!t]
    \centering
    \includegraphics[width=\linewidth]{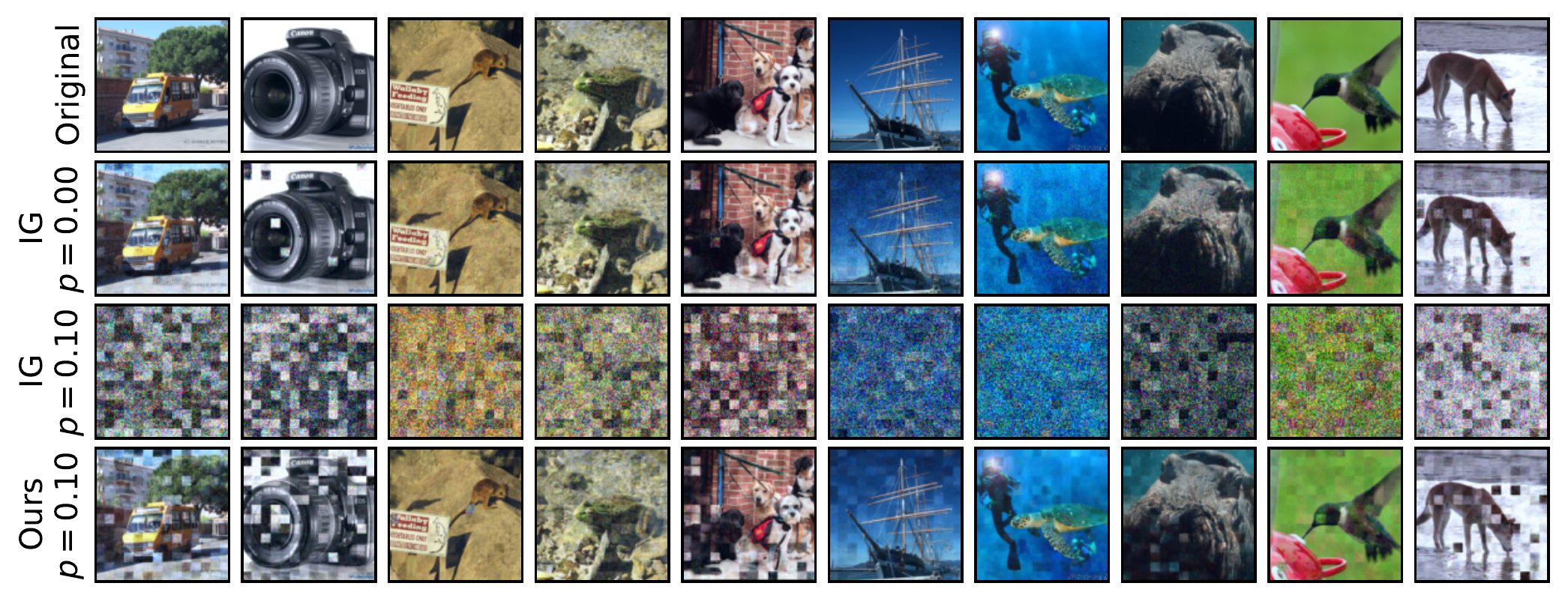}
    \caption{
    Example reconstructions for batchsize $\mathcal{B}=1$ for ViT-B/16 on ImageNet.
    }
    \label{fig:apx_in_vit16}
\end{figure*}

\begin{figure*}[!t]
    \centering
    \includegraphics[width=\linewidth]{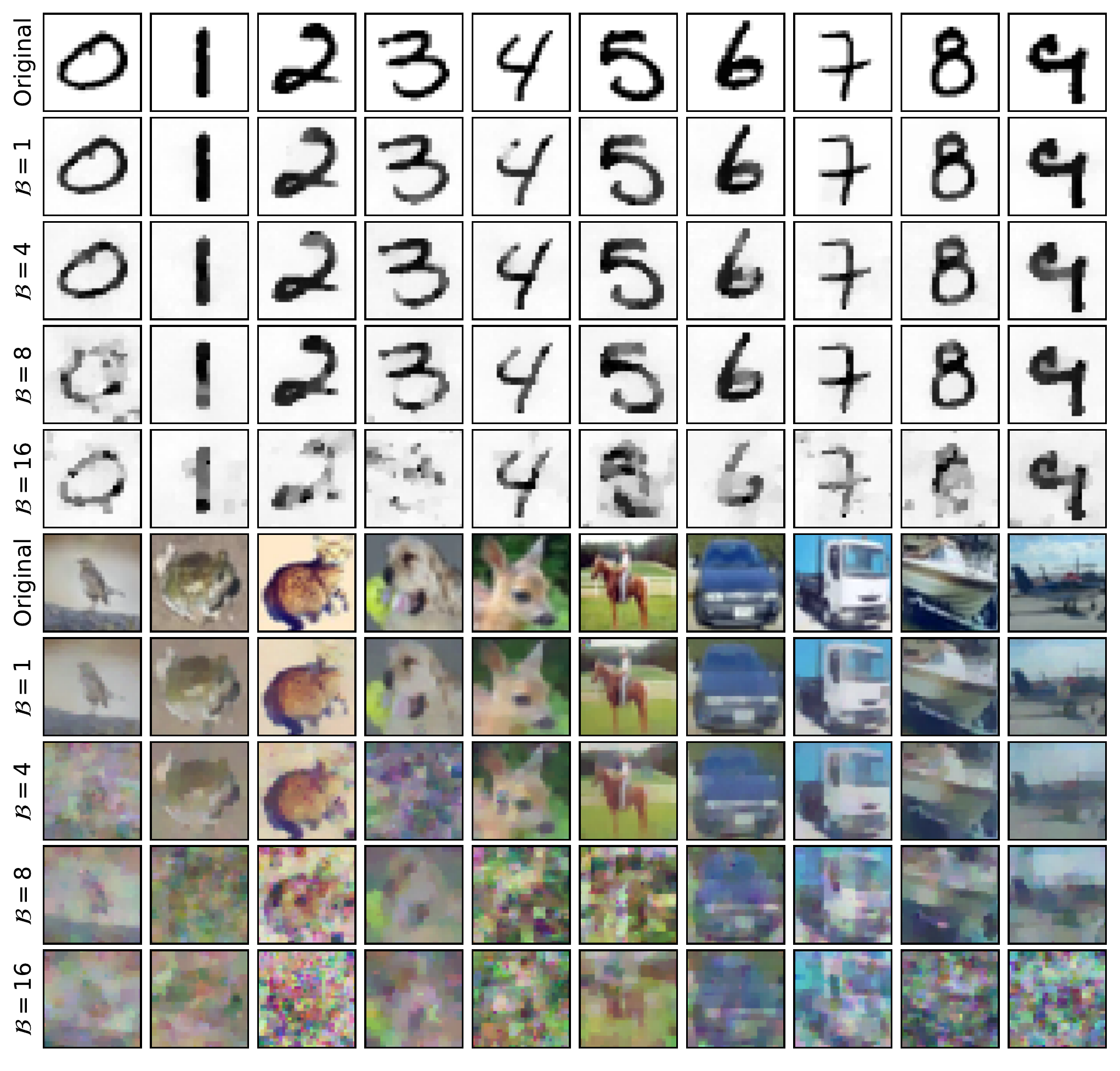}
    \caption{
    Example DIA reconstructions (ours) for increasing batchsizes $\mathcal{B}$ and fixed dropout rate $p=0.25$ for ViT on MNIST and CIFAR-10.
    }
    \label{fig:apx_vit_bs}
\end{figure*}

More detailed results for the experiments in section 5 of our paper are reported in Tab.~\ref{tab:IGA}-\ref{tab:FullOtherModels}.
Fig.~\ref{fig:apx_mnist_mlp}-\ref{fig:apx_vit_bs} display more example reconstructions.

\subsection{Dropout Mask Initialization}
\label{sec:mask_init}

\subsubsection{Analytical Mask Reconstruction}
In certain cases the dropout masks $\Psi_C$ for fully-connected layers can be reconstructed analytically, because the corresponding positions of the gradients become zero.
Let $\psi_C^{(i)}$ be the dropout mask which the client applies to the feature output of a fully-connected layer $i$ with $n$ neurons, i.e. $\psi_C^{(i)}\in \lbrace 0,1 \rbrace^{(\mathcal{B}\times n)}$.
In the case of batchsize $\mathcal{B} = 1$ and non-ReLU activation functions, the dropout mask $\psi_C^{(i)}$ can be calculated from the gradient $\nabla_C^{(i)}\in\mathbb{R}^{(n\times m)}$ of the corresponding fully-connected layer as follows:
\begin{equation}
    \label{eq:det_mask_init}
    \psi_C^{(i)}\lbrack k \rbrack = 
    \begin{cases}
      1, & \text{if}\ \sum_{j=1}^{m} | \nabla_C^{(i)} \lbrack k,j \rbrack | > 0 \\
      0, & \text{otherwise}
    \end{cases}
\end{equation}
In Eq.~\ref{eq:det_mask_init}, $\psi_C^{(i)}\lbrack k \rbrack$ refers to the $k$th element of the clients dropout mask which is multiplied to the output of the $k$th neuron in the corresponding fully-connected layer $i$.
This type of mask reconstruction can eliminate the inaccurate estimation of some masks and ensure that $\psi_A^{(i)} = \psi_C^{(i)}$.
However, the analytical reconstruction of dropout masks is only applicable in specific cases, i.e. batchsize $\mathcal{B} = 1$ and non-ReLU activation functions for fully-connected layers.
We performed preliminary experiments and found such analytical mask reconstructions have no benefit regarding reconstruction quality compared to iterative optimization.
Results can be found in Tab.~\ref{tab:mask_init}.

\subsubsection{Other Mask Initializations}

Consistent with recent work~\cite{Geiping2020, Enthoven2020a, Yin2021, Kaissis2021, Jin2021, Scheliga2022a, Scheliga2022b, Zhang2022, Gupta2022} our threat model assumes the attacker to have knowledge of the model architecture and thus the dropout rate $p$.
The attacker uses this knowledge to initialize the dropout masks $\psi_A^{(i)}$ by sampling random Bernoulli variables $\text{Bernoulli}(p)$ (c.f. Algorithm \ref{alg:dia} line 1).
In a set of preliminary experiments we also tried initializing $\psi_A^{(i)}$ from $\mathcal{N}(1-p, \frac{1}{\sqrt{n_i}})$, where $n_i$ is the size of dropout mask $\psi_A^{(i)}$.

In case the attacker has no information on the dropout rate $p$, these initialization schemes are obsolete.
Instead, the attacker would have to guess $p$.
Therefore we also tried initializing $\psi_A^{(i)}$ from $\mathcal{N}(0.5, 0.25)$ for varying dropout rates $p$.
We found that the type of initialization has no significant impact on the reconstruction quality for various dropout rates $p$.
Results can be found in Tab.~\ref{tab:mask_init}.
Please note that our proposed regularization method cannot be used if $p$ is unknown to the attacker, which would negatively impact the reconstruction quality as observed in Fig.~\ref{fig:DIAheatmaps} a) and c).

\begin{table*}
\centering
\resizebox{\linewidth}{!}{%
\begin{tabular}{c|c|c|c|c|c|c|c|c|c}
\toprule
 Model & $p$ & Initialization & SSIM $\uparrow$ & PSNR $\uparrow$ & MSE $\downarrow$ & LPIPS $\downarrow$ & MMD & TRD & Parameters \\
\midrule
\multirow[c]{12}{*}{MLP} & \multirow[c]{4}{*}{0.25} & Analytical & {\cellcolor[HTML]{0A7B41}} \color[HTML]{F1F1F1} \bfseries 0.96 ($\pm$0.04) & {\cellcolor[HTML]{006837}} \color[HTML]{F1F1F1} \bfseries 31.91 ($\pm$4.89) & {\cellcolor[HTML]{05713C}} \color[HTML]{F1F1F1} \bfseries 0.03 ($\pm$0.06) & {\cellcolor[HTML]{0C7F43}} \color[HTML]{F1F1F1} \bfseries 0.03 ($\pm$0.04) & \bfseries 0.02 ($\pm$0.02) & \bfseries 0.01 ($\pm$0.01) & \bfseries 4608 \\
 &  & Bernoulli($p$) & {\cellcolor[HTML]{0C7F43}} \color[HTML]{F1F1F1} 0.95 ($\pm$0.04) & {\cellcolor[HTML]{06733D}} \color[HTML]{F1F1F1} 31.06 ($\pm$4.93) & {\cellcolor[HTML]{07753E}} \color[HTML]{F1F1F1} 0.04 ($\pm$0.06) & {\cellcolor[HTML]{108647}} \color[HTML]{F1F1F1} 0.04 ($\pm$0.04) & 0.03 ($\pm$0.01) & \bfseries 0.01 ($\pm$0.01) & \bfseries 4608 \\
 &  & $\mathcal{N}(1-p, \frac{1}{\sqrt{n_i}})$ & {\cellcolor[HTML]{0C7F43}} \color[HTML]{F1F1F1} 0.95 ($\pm$0.03) & {\cellcolor[HTML]{07753E}} \color[HTML]{F1F1F1} 30.92 ($\pm$4.50) & {\cellcolor[HTML]{07753E}} \color[HTML]{F1F1F1} 0.04 ($\pm$0.06) & {\cellcolor[HTML]{0C7F43}} \color[HTML]{F1F1F1} \bfseries 0.03 ($\pm$0.03) & 0.03 ($\pm$0.01) & \bfseries 0.01 ($\pm$0.01) & \bfseries 4608 \\
 &  & $\mathcal{N}(0.5, 0.25)$ & {\cellcolor[HTML]{0C7F43}} \color[HTML]{F1F1F1} 0.95 ($\pm$0.04) & {\cellcolor[HTML]{0C7F43}} \color[HTML]{F1F1F1} 30.39 ($\pm$3.67) & {\cellcolor[HTML]{07753E}} \color[HTML]{F1F1F1} 0.04 ($\pm$0.06) & {\cellcolor[HTML]{0C7F43}} \color[HTML]{F1F1F1} \bfseries 0.03 ($\pm$0.04) & 0.03 ($\pm$0.01) & \bfseries 0.01 ($\pm$0.01) & \bfseries 4608 \\
\bottomrule
\end{tabular}
}
\caption{
Reconstruction quality metrics computed from gradients attacked with DIA with different mask initialization schemes for MLP and ViT on CIFAR-10 for increasing dropout rates $p$ and batchsize $\mathcal{B}=1$.
Arrows indicate direction of improvement.
Bold and italic formatting highlight best and worst results respectively.
}
\label{tab:mask_init}
\end{table*}

\begin{table*}[t]
\centering
\resizebox{\linewidth}{!}{%
\begin{tabular}{c|c|c|c|c|c|c|c|c}
\toprule
$p$ & Attack &  SSIM $\uparrow$ & PSNR $\uparrow$ & MSE $\downarrow$ & LPIPS $\downarrow$ & MMD & TRD & Parameters \\
\midrule
\multirow[c]{2}{*}{0.00} & IG & {\cellcolor[HTML]{36A657}} \color[HTML]{F1F1F1} 0.86 ($\pm$0.07) & {\cellcolor[HTML]{A5D86A}} \color[HTML]{000000} 25.98 ($\pm$2.14) & {\cellcolor[HTML]{016A38}} \color[HTML]{F1F1F1} 0.02 ($\pm$0.00) & {\cellcolor[HTML]{5AB760}} \color[HTML]{F1F1F1} 0.13 ($\pm$0.06) & - & - & \bfseries 3072 \\
 & APRIL & {\cellcolor[HTML]{07753E}} \color[HTML]{F1F1F1} \bfseries 0.97 ($\pm$0.04) & {\cellcolor[HTML]{006837}} \color[HTML]{F1F1F1} \bfseries 37.12 ($\pm$3.65) & {\cellcolor[HTML]{006837}} \color[HTML]{F1F1F1} \bfseries 0.00 ($\pm$0.00) & {\cellcolor[HTML]{07753E}} \color[HTML]{F1F1F1} \bfseries 0.02 ($\pm$0.02) & - & - & \bfseries 3072 \\
 \cline{1-9}
\multirow[c]{4}{*}{0.10} & IG & {\cellcolor[HTML]{B30D26}} \color[HTML]{F1F1F1} 0.03 ($\pm$0.02) & {\cellcolor[HTML]{F46D43}} \color[HTML]{F1F1F1} 7.43 ($\pm$0.55) & {\cellcolor[HTML]{F57547}} \color[HTML]{F1F1F1} 2.84 ($\pm$0.44) & {\cellcolor[HTML]{AB0626}} \color[HTML]{F1F1F1} 0.69 ($\pm$0.06) & - & - & \bfseries 3072 \\
 & APRIL & {\cellcolor[HTML]{B30D26}} \color[HTML]{F1F1F1} 0.03 ($\pm$0.02) & {\cellcolor[HTML]{F46D43}} \color[HTML]{F1F1F1} 7.41 ($\pm$0.53) & {\cellcolor[HTML]{F57245}} \color[HTML]{F1F1F1} 2.85 ($\pm$0.42) & {\cellcolor[HTML]{AB0626}} \color[HTML]{F1F1F1} 0.69 ($\pm$0.06) & - & - & \bfseries 3072 \\
 & DIA-IG & {\cellcolor[HTML]{57B65F}} \color[HTML]{F1F1F1} 0.82 ($\pm$0.06) & {\cellcolor[HTML]{DAF08D}} \color[HTML]{000000} 22.09 ($\pm$1.59) & {\cellcolor[HTML]{05713C}} \color[HTML]{F1F1F1} 0.08 ($\pm$0.05) & {\cellcolor[HTML]{82C966}} \color[HTML]{000000} 0.17 ($\pm$0.05) & \bfseries 0.23 ($\pm$0.02) & 0.15 ($\pm$0.02) & \itshape 489792 \\
\bottomrule
\end{tabular}
}
\caption{
Reconstruction quality metrics computed from gradients attacked with IG and iterative APRIL as well as the corresponding DIA versions (ours) for ViT on CIFAR-10 for increasing dropout rates $p$ and batchsize $\mathcal{B}=1$.
Arrows indicate direction of improvement.
Bold and italic formatting highlight best and worst results respectively.
}
\label{tab:APRIL}
\end{table*}

\begin{table*}
\centering
% \resizebox{0.9\linewidth}{!}{%
\begin{tabular}{c|c|c|c|c|c|c|c}
\toprule
 & Model & $p$ & SSIM $\uparrow$ & PSNR $\uparrow$ & MSE $\downarrow$ & LPIPS $\downarrow$ & Parameters \\
\midrule
\multirow[c]{8}{*}{\rotatebox{90}{MNIST}} & \multirow[c]{4}{*}{MLP} & 0.00& {\cellcolor[HTML]{006837}} \color[HTML]{F1F1F1} \bfseries 1.00 ($\pm$0.00) & {\cellcolor[HTML]{006837}} \color[HTML]{F1F1F1} \bfseries 57.73 ($\pm$1.40) & {\cellcolor[HTML]{006837}} \color[HTML]{F1F1F1} \bfseries 0.00 ($\pm$0.00) & {\cellcolor[HTML]{006837}} \color[HTML]{F1F1F1} \bfseries 0.00 ($\pm$0.00) & \bfseries \itshape 784 \\
 &  & 0.25& {\cellcolor[HTML]{6BBF64}} \color[HTML]{000000} 0.79 ($\pm$0.07) & {\cellcolor[HTML]{FDB365}} \color[HTML]{000000} 18.03 ($\pm$1.61) & {\cellcolor[HTML]{138C4A}} \color[HTML]{F1F1F1} 0.18 ($\pm$0.06) & {\cellcolor[HTML]{FFFEBE}} \color[HTML]{000000} 0.33 ($\pm$0.05) & \bfseries \itshape 784 \\
 &  & 0.50& {\cellcolor[HTML]{DCF08F}} \color[HTML]{000000} 0.59 ($\pm$0.09) & {\cellcolor[HTML]{F36B42}} \color[HTML]{F1F1F1} 11.48 ($\pm$2.04) & {\cellcolor[HTML]{BDE379}} \color[HTML]{000000} 0.83 ($\pm$0.36) & {\cellcolor[HTML]{EE613E}} \color[HTML]{F1F1F1} 0.54 ($\pm$0.06) & \bfseries \itshape 784 \\
 &  & 0.75& {\cellcolor[HTML]{FDAD60}} \color[HTML]{000000} 0.30 ($\pm$0.13) & {\cellcolor[HTML]{DB382B}} \color[HTML]{F1F1F1} 6.59 ($\pm$1.27) & {\cellcolor[HTML]{A50026}} \color[HTML]{F1F1F1} \itshape 2.40 ($\pm$0.59) & {\cellcolor[HTML]{A50026}} \color[HTML]{F1F1F1} \itshape 0.66 ($\pm$0.04) & \bfseries \itshape 784 \\
\cline{2-8}
 & \multirow[c]{4}{*}{ViT} & 0.00& {\cellcolor[HTML]{05713C}} \color[HTML]{F1F1F1} 0.98 ($\pm$0.01) & {\cellcolor[HTML]{FFF6B0}} \color[HTML]{000000} 27.17 ($\pm$1.28) & {\cellcolor[HTML]{026C39}} \color[HTML]{F1F1F1} 0.02 ($\pm$0.01) & {\cellcolor[HTML]{07753E}} \color[HTML]{F1F1F1} 0.02 ($\pm$0.01) & \bfseries \itshape 784 \\
\bottomrule
\end{tabular}
% }
\caption{
Reconstruction quality metrics computed from gradients attacked with IG for MLP and ViT on MNIST and CIFAR-10 for increasing dropout rates $p$ and batchsize $\mathcal{B}=1$.
Arrows indicate direction of improvement.
Bold and italic formatting highlight best and worst results respectively.
}
\label{tab:IGA}
\end{table*}

\begin{table*}
\centering
% \resizebox{0.9\linewidth}{!}{%
\begin{tabular}{c|c|c|c|c|c|c|c}
\toprule
 & Model & $p$ & SSIM $\uparrow$ & PSNR $\uparrow$ & MSE $\downarrow$ & LPIPS $\downarrow$ & Parameters \\
\midrule
\multirow[c]{6}{*}{\rotatebox{90}{MNIST}} & \multirow[c]{3}{*}{MLP} & 0.25& {\cellcolor[HTML]{006837}} \color[HTML]{F1F1F1} \bfseries 1.00 ($\pm$0.00) & {\cellcolor[HTML]{006837}} \color[HTML]{F1F1F1} \bfseries 57.45 ($\pm$1.58) & {\cellcolor[HTML]{006837}} \color[HTML]{F1F1F1} \bfseries 0.00 ($\pm$0.00) & {\cellcolor[HTML]{006837}} \color[HTML]{F1F1F1} \bfseries 0.00 ($\pm$0.00) & \bfseries \itshape 784 \\
 &  & 0.50& {\cellcolor[HTML]{006837}} \color[HTML]{F1F1F1} \bfseries 1.00 ($\pm$0.02) & {\cellcolor[HTML]{08773F}} \color[HTML]{F1F1F1} 55.58 ($\pm$7.20) & {\cellcolor[HTML]{006837}} \color[HTML]{F1F1F1} \bfseries 0.00 ($\pm$0.01) & {\cellcolor[HTML]{08773F}} \color[HTML]{F1F1F1} 0.01 ($\pm$0.03) & \bfseries \itshape 784 \\
 &  & 0.75& {\cellcolor[HTML]{57B65F}} \color[HTML]{F1F1F1} \itshape 0.82 ($\pm$0.09) & {\cellcolor[HTML]{FDC372}} \color[HTML]{000000} \itshape 19.53 ($\pm$2.35) & {\cellcolor[HTML]{A50026}} \color[HTML]{F1F1F1} \itshape 0.13 ($\pm$0.07) & {\cellcolor[HTML]{A50026}} \color[HTML]{F1F1F1} \itshape 0.29 ($\pm$0.08) & \bfseries \itshape 784 \\
\cline{2-8}
 & \multirow[c]{3}{*}{ViT} & 0.25& {\cellcolor[HTML]{026C39}} \color[HTML]{F1F1F1} 0.99 ($\pm$0.00) & {\cellcolor[HTML]{E2F397}} \color[HTML]{000000} 33.13 ($\pm$1.15) & {\cellcolor[HTML]{138C4A}} \color[HTML]{F1F1F1} 0.01 ($\pm$0.00) & {\cellcolor[HTML]{08773F}} \color[HTML]{F1F1F1} 0.01 ($\pm$0.01) & \bfseries \itshape 784 \\
 &  & 0.50& {\cellcolor[HTML]{026C39}} \color[HTML]{F1F1F1} 0.99 ($\pm$0.00) & {\cellcolor[HTML]{CDEA83}} \color[HTML]{000000} 35.88 ($\pm$1.53) & {\cellcolor[HTML]{006837}} \color[HTML]{F1F1F1} \bfseries 0.00 ($\pm$0.00) & {\cellcolor[HTML]{08773F}} \color[HTML]{F1F1F1} 0.01 ($\pm$0.01) & \bfseries \itshape 784 \\
\bottomrule
\end{tabular}
% }
\caption{
Reconstruction quality metrics computed from gradients attacked by a well-informed attacker (WIIG) that has knowledge of the clients' dropout masks $\Psi_C$.
Results are reported for MLP and ViT on MNIST and CIFAR-10 for increasing dropout rates $p$ and batchsize $\mathcal{B}=1$.
Arrows indicate direction of improvement.
Bold and italic formatting highlight best and worst results respectively.
}
\label{tab:PowerMasking}
\end{table*}

\begin{table*}[t]
\centering
\label{tab:DIA}
\resizebox{\linewidth}{!}{%
\begin{tabular}{c|c|c|c|c|c|c|c|c|c|c}
\toprule
 & Model & $p$ & $\mathcal{B}$ & SSIM $\uparrow$ & PSNR $\uparrow$ & MSE $\downarrow$ & LPIPS $\downarrow$ & MMD & TRD & Parameters \\
\midrule
\multirow[c]{24}{*}{\rotatebox{90}{MNIST}} & \multirow[c]{12}{*}{MLP} & \multirow[c]{4}{*}{0.25} & 1 & {\cellcolor[HTML]{006837}} \color[HTML]{F1F1F1} \bfseries 1.00 ($\pm$0.00) & {\cellcolor[HTML]{006837}} \color[HTML]{F1F1F1} \bfseries 48.05 ($\pm$6.25) & {\cellcolor[HTML]{006837}} \color[HTML]{F1F1F1} \bfseries 0.00 ($\pm$0.00) & {\cellcolor[HTML]{006837}} \color[HTML]{F1F1F1} \bfseries 0.00 ($\pm$0.01) & \bfseries 0.01 ($\pm$0.01) & 0.01 ($\pm$0.01) & \bfseries 2320 \\
 &  &  & 4 & {\cellcolor[HTML]{006837}} \color[HTML]{F1F1F1} \bfseries 1.00 ($\pm$0.00) & {\cellcolor[HTML]{07753E}} \color[HTML]{F1F1F1} 46.59 ($\pm$5.63) & {\cellcolor[HTML]{006837}} \color[HTML]{F1F1F1} \bfseries 0.00 ($\pm$0.00) & {\cellcolor[HTML]{006837}} \color[HTML]{F1F1F1} \bfseries 0.00 ($\pm$0.01) & 0.05 ($\pm$0.07) & 0.01 ($\pm$0.00) & 9280 \\
 &  &  & 8 & {\cellcolor[HTML]{006837}} \color[HTML]{F1F1F1} \bfseries 1.00 ($\pm$0.00) & {\cellcolor[HTML]{05713C}} \color[HTML]{F1F1F1} 46.98 ($\pm$4.71) & {\cellcolor[HTML]{006837}} \color[HTML]{F1F1F1} \bfseries 0.00 ($\pm$0.00) & {\cellcolor[HTML]{006837}} \color[HTML]{F1F1F1} \bfseries 0.00 ($\pm$0.01) & 0.14 ($\pm$0.09) & 0.01 ($\pm$0.00) & 18560 \\
 &  &  & 16 & {\cellcolor[HTML]{006837}} \color[HTML]{F1F1F1} \bfseries 1.00 ($\pm$0.00) & {\cellcolor[HTML]{097940}} \color[HTML]{F1F1F1} 46.24 ($\pm$3.94) & {\cellcolor[HTML]{006837}} \color[HTML]{F1F1F1} \bfseries 0.00 ($\pm$0.00) & {\cellcolor[HTML]{006837}} \color[HTML]{F1F1F1} \bfseries 0.00 ($\pm$0.01) & 0.22 ($\pm$0.06) & 0.01 ($\pm$0.00) & 37120 \\
\bottomrule
\end{tabular}
}
\caption{
Reconstruction quality metrics computed from gradients attacked with DIA (ours) for MLP and ViT on MNIST and CIFAR-10 for increasing dropout rates $p$ and batchsizes $\mathcal{B}$.
Arrows indicate direction of improvement.
Bold and italic formatting highlight best and worst results respectively.
}
\end{table*}

\begin{table*}[t]
\centering
\resizebox{\linewidth}{!}{%
\begin{tabular}{c|c|c|c|c|c|c|c|c|c|c}
\toprule
 & Model & $\lambda_{\text{mask}}$ & $\mathcal{B}$ & SSIM $\uparrow$ & PSNR $\uparrow$ & MSE $\downarrow$ & LPIPS $\downarrow$ & MMD & TRD & Parameters \\
\midrule
\multirow[c]{32}{*}{\rotatebox{90}{MNIST}} & \multirow[c]{16}{*}{MLP} & \multirow[c]{4}{*}{0} & 1 & {\cellcolor[HTML]{006837}} \color[HTML]{F1F1F1} \bfseries 1.00 ($\pm$0.00) & {\cellcolor[HTML]{006837}} \color[HTML]{F1F1F1} 48.05 ($\pm$6.25) & {\cellcolor[HTML]{006837}} \color[HTML]{F1F1F1} \bfseries 0.00 ($\pm$0.00) & {\cellcolor[HTML]{006837}} \color[HTML]{F1F1F1} \bfseries 0.00 ($\pm$0.01) & \bfseries 0.01 ($\pm$0.01) & 0.01 ($\pm$0.01) & \bfseries 2320 \\
 &  &  & 4 & {\cellcolor[HTML]{006837}} \color[HTML]{F1F1F1} \bfseries 1.00 ($\pm$0.00) & {\cellcolor[HTML]{08773F}} \color[HTML]{F1F1F1} 46.59 ($\pm$5.63) & {\cellcolor[HTML]{006837}} \color[HTML]{F1F1F1} \bfseries 0.00 ($\pm$0.00) & {\cellcolor[HTML]{006837}} \color[HTML]{F1F1F1} \bfseries 0.00 ($\pm$0.01) & 0.05 ($\pm$0.07) & 0.01 ($\pm$0.00) & 9280 \\
 &  &  & 8 & {\cellcolor[HTML]{006837}} \color[HTML]{F1F1F1} \bfseries 1.00 ($\pm$0.00) & {\cellcolor[HTML]{06733D}} \color[HTML]{F1F1F1} 46.98 ($\pm$4.71) & {\cellcolor[HTML]{006837}} \color[HTML]{F1F1F1} \bfseries 0.00 ($\pm$0.00) & {\cellcolor[HTML]{006837}} \color[HTML]{F1F1F1} \bfseries 0.00 ($\pm$0.01) & 0.14 ($\pm$0.09) & 0.01 ($\pm$0.00) & 18560 \\
 &  &  & 16 & {\cellcolor[HTML]{006837}} \color[HTML]{F1F1F1} \bfseries 1.00 ($\pm$0.00) & {\cellcolor[HTML]{0A7B41}} \color[HTML]{F1F1F1} 46.24 ($\pm$3.94) & {\cellcolor[HTML]{006837}} \color[HTML]{F1F1F1} \bfseries 0.00 ($\pm$0.00) & {\cellcolor[HTML]{006837}} \color[HTML]{F1F1F1} \bfseries 0.00 ($\pm$0.01) & 0.22 ($\pm$0.06) & 0.01 ($\pm$0.00) & 37120 \\
\bottomrule
\end{tabular}
}
\caption{
Reconstruction quality metrics computed from gradients attacked with DIA (ours) for MLP and ViT on MNIST and CIFAR-10 for increasing regularization term weights $\lambda_{\text{mask}}$ and batchsizes $\mathcal{B}$.
Dropout rate is fixed to $p=0.25$.
Arrows indicate direction of improvement.
Bold and italic formatting highlight best and worst results respectively.
}
\label{tab:DIAreg}
\end{table*}

\begin{table*}
\centering
\resizebox{\linewidth}{!}{%
\begin{tabular}{c|c|c|c|c|c|c|c|c|c|c}
\toprule
 & Model & $p$ & Attack & SSIM $\uparrow$ & PSNR $\uparrow$ & MSE $\downarrow$ & LPIPS $\downarrow$ & MMD & TRD & Parameters \\
\midrule
\multirow[c]{3}{*}{\rotatebox{90}{ImgNet}} & \multirow[c]{3}{*}{ViT-B/16} & 0.00& IG & {\cellcolor[HTML]{E8F59F}} \color[HTML]{000000} 0.56 ($\pm$0.14) & {\cellcolor[HTML]{006837}} \color[HTML]{F1F1F1} \bfseries 21.04 ($\pm$2.53) & {\cellcolor[HTML]{0D8044}} \color[HTML]{F1F1F1} \bfseries 0.20 ($\pm$0.32) & {\cellcolor[HTML]{F1F9AC}} \color[HTML]{000000} 0.37 ($\pm$0.09) & - & - & \bfseries 150528 \\
\cline{3-11}
 &  & \multirow[c]{2}{*}{0.10} & IG & {\cellcolor[HTML]{A90426}} \color[HTML]{F1F1F1} \itshape 0.01 ($\pm$0.00) & {\cellcolor[HTML]{FED27F}} \color[HTML]{000000} \itshape 7.84 ($\pm$0.77) & {\cellcolor[HTML]{A50026}} \color[HTML]{F1F1F1} \itshape 3.67 ($\pm$0.71) & {\cellcolor[HTML]{A50026}} \color[HTML]{F1F1F1} \itshape 0.80 ($\pm$0.04) & - & - & \bfseries 150528 \\
 &  &  & DIA & {\cellcolor[HTML]{98D368}} \color[HTML]{000000} \bfseries 0.72 ($\pm$0.11) & {\cellcolor[HTML]{108647}} \color[HTML]{F1F1F1} 19.72 ($\pm$5.16) & {\cellcolor[HTML]{249D53}} \color[HTML]{F1F1F1} 0.43 ($\pm$0.59) & {\cellcolor[HTML]{CBE982}} \color[HTML]{000000} \bfseries 0.30 ($\pm$0.11) & \bfseries \itshape 0.13 ($\pm$0.01) & \bfseries \itshape 0.02 ($\pm$0.01) & \itshape 16783632 \\
\midrule
\multirow[c]{14}{*}{\rotatebox{90}{MNIST}} & \multirow[c]{7}{*}{LeNet} & 0.00& IG & {\cellcolor[HTML]{0C7F43}} \color[HTML]{F1F1F1} \bfseries 0.95 ($\pm$0.03) & {\cellcolor[HTML]{006837}} \color[HTML]{F1F1F1} \bfseries 24.60 ($\pm$3.55) & {\cellcolor[HTML]{108647}} \color[HTML]{F1F1F1} \bfseries 0.09 ($\pm$0.05) & {\cellcolor[HTML]{08773F}} \color[HTML]{F1F1F1} \bfseries 0.02 ($\pm$0.01) & - & - & \bfseries 2352 \\
\bottomrule
\end{tabular}
}
\caption{
Reconstruction quality metrics computed from gradients attacked with IG and DIA (ours) for ViT-B/16 on ImageNet as well as LeNet and ResNet on MNIST and CIFAR-10 for increasing dropout rates $p$ and batchsize $\mathcal{B}=1$.
Arrows indicate direction of improvement.
Bold and italic formatting highlight best and worst results respectively.
}
\label{tab:FullOtherModels}
\end{table*}

\clearpage

\end{document}